\documentclass[sigconf,screen]{acmart}

\usepackage{subfig}
\usepackage[normalem]{ulem}
\usepackage{xspace}
\newcommand{\projectname}{SoD$^2$\xspace}

\newcommand{\revisioncr}[1]{\textcolor{black}{#1}} %
\usepackage[showboxes,overlay]{textpos}
\TPMargin*{3pt}
\textblockcolour{cyan}

\newcommand{\opset}[1]{\tt #1}
\usepackage{hyperref}
\usepackage{soul}
\usepackage{ctable}
\usepackage{booktabs}
\usepackage{longtable}
\usepackage{float}
\usepackage{lscape}
\usepackage[T1]{fontenc}
\usepackage{algcompatible}
\usepackage{amsmath} %
\usepackage{amsfonts}
\usepackage{pifont}
\usepackage{balance}
\usepackage{array}
\newcolumntype{P}[1]{>{\centering\arraybackslash}p{#1}}
\usepackage{listings}
\definecolor{codegray}{RGB}{0,92,240}
\definecolor{codepurple}{rgb}{0.58,0,0.82}
\definecolor{keyword}{RGB}{186, 45, 162}
\lstdefinelanguage{mylanguage}{
  keywords={True, False, return, switch, if, in, while, do, else, case, break, def, for},
  keywordstyle=\color{blue}\bfseries,
  ndkeywords={class, export, boolean, throw, implements, import, this},
  ndkeywordstyle=\color{darkgray}\bfseries,
  identifierstyle=\color{black},
  sensitive=false,
  comment=[l]{\#},
  morecomment=[s]{/*}{*/},
  commentstyle=\color{purple}\ttfamily,
  stringstyle=\color{red}\ttfamily,
  morestring=[b]',
  morestring=[b]"
}
\lstdefinestyle{mystyle}{
    language=mylanguage,
    backgroundcolor=\color{white},
    numberstyle=\tiny\color{black},
    basicstyle=\ttfamily\footnotesize,
    breakatwhitespace=false,
    breaklines=true,
    captionpos=b,
    keepspaces=true,
    numbers=left,
    numbersep=3pt,
    stepnumber=1,
    showspaces=false,
    showstringspaces=false,
    showtabs=false,
    tabsize=2,
    morekeywords={entry},
    frame=tb,
}
\usepackage[ruled,vlined,linesnumbered]{algorithm2e}
\usepackage{mathptmx}
\usepackage[normalem]{ulem}
\usepackage{multirow}
\usepackage{multicol}
\usepackage{makecell}
\usepackage{xspace}
\usepackage{xcolor}
\usepackage{enumitem}
\usepackage{lipsum}
\usepackage{ragged2e}

\copyrightyear{2024} 
\acmYear{2024} 
\setcopyright{acmlicensed}\acmConference[ASPLOS '24]{29th ACM International Conference on Architectural Support for Programming Languages and Operating Systems, Volume 1}{April 27-May 1, 2024}{La Jolla, CA, USA}
\acmBooktitle{29th ACM International Conference on Architectural Support for Programming Languages and Operating Systems, Volume 1 (ASPLOS '24), April 27-May 1, 2024, La Jolla, CA, USA}
\acmPrice{15.00}
\acmDOI{10.1145/3617232.3624869}
\acmISBN{979-8-4007-0372-0/24/04}

\begin{document}

\title{\projectname: Statically Optimizing Dynamic Deep Neural Network Execution}

\author{Wei Niu}
\email{wniu@uga.edu}
\affiliation{
  \institution{University of Georgia}
  \city{Athens}
  \state{GA}
  \country{USA}
}
\authornote{This work was primarily done while the author was at William \& Mary.}

\author{Gagan Agrawal}
\email{gagrawal@uga.edu}
\affiliation{
  \institution{University of Georgia}
  \state{Athens}
  \city{GA}
  \country{USA}
}

\author{Bin Ren}
\email{bren@wm.edu}
\affiliation{
  \institution{William \& Mary}
  \city{Williamsburg}
  \state{VA}
  \country{USA}
}

\renewcommand{\shortauthors}{Niu et al.}

\begin{abstract}
Though many compilation and runtime systems have been developed for DNNs in
recent years, the focus has largely been on {\em static DNNs}. Dynamic DNNs,
where tensor shapes and sizes and even the set of operators used are dependent
upon the input and/or execution, are becoming common. This paper presents \projectname, a comprehensive framework for optimizing Dynamic DNNs. The basis
of our approach is a classification of common operators that form DNNs, and
the use of this classification towards a Rank and Dimension Propagation (RDP) 
method. This framework statically determines the shapes of operators
as known constants, symbolic constants, or operations on these.
Next, using RDP we enable a series of optimizations, like fused code
generation, execution (order) planning, and even runtime memory allocation
plan generation. By evaluating the framework on 10 emerging Dynamic DNNs and comparing it against several existing systems, we demonstrate both reductions in execution
latency and memory requirements, with RDP-enabled key optimizations responsible
for much of the gains. 
Our evaluation results show that \projectname runs up to $3.9\times$ faster than these systems while saving up to $88\%$ peak memory consumption.

\end{abstract}

\ccsdesc[500]{Computing methodologies~Neural networks}
\ccsdesc[500]{Software and its engineering~Source code generation}
\ccsdesc[500]{Human-centered computing~Mobile computing}

\keywords{dynamic neural network, compiler optimization, mobile device}

\maketitle

\section{Introduction} 

Deep Neural Networks are enabling several of the most exciting and innovative 
applications that are executed on a variety of computing devices, 
ranging from servers to edge and mobile devices. From a systems research viewpoint, this had led to a large set of ongoing  projects  on optimizing DNN inference (and training) tasks~\cite{TensorFlow-Lite,han2016mcdnn,lane2016deepx,yao2017deepsense,huynh2017deepmon,vasilache2018tensor,xu2018deepcache,lee2019mobisr,Ali-MNN} 
as well as   tensor compilers~\cite{halide,kjolstad2017tensor,lattner2021mlir}.  

Most of the work on optimizing DNNs considers {\em static models} that   
are characterized by the following two properties: 1) input and output shapes and sizes for each layer are known a prior,  
and 2) the execution path is fixed, i.e., independent of the input.   
In {\em dynamic models}, in contrast,  one or both of the above two properties are no longer true, 
and such models are now becoming prevalent. 
For example,  
 Skipnet~\cite{wang2018skipnet} decides,  based on the input, whether to include or exclude certain operators (or layers). A different form 
 of dynamism seen in transformers for NLP like BERT~\cite{devlin2018bert} or cutting-edge computer vision models~\cite{rao2021dynamicvit,rombach2022high,kirillov2023segany}  can take inputs with different shapes and/or apply variable portions of filter kernels during the execution. 
\revisioncr{
Consider a commonly used dataset like Wikipedia. The length of input sequences typically varies from 32 to 512~\cite{zeng2022boosting}, creating significant dynamism in text processing. Similarly, neural networks for image/video processing often deal with images/videos of varying resolutions that dynamically change based on network conditions and player settings.
}
At least three factors have contributed to the popularity 
of dynamic  models and this trend is expected to continue: the need  for  adapting to  computational capacities 
of different devices, the need for supporting different types of input (e.g. images of different resolutions), 
and the need for achieving high accuracy for different scenarios. 

Dynamic shapes, sizes, and control flow in these models pose 
many challenges for the optimizations that have been key to 
obtaining high efficiency. For example, loop fusion~\cite{tensorflow-xla,tensorflow-grappler, rammer-osdi20,niu2021dnnfusion}
cannot be applied ~\cite{zhu2021disc, shen2021nimble, zheng2022dietcode} if  we do not know that the index space of two 
loops (which likely is the same as the dimensions of respective input tensors)  
is identical. Planning the execution order~\cite{ahn2020ordering} to reduce memory requirements or otherwise planning memory allocation~\cite{pisarchyk2020efficient} is, similarly, not possible if tensor sizes are not statically known. 

While many of the existing systems for DNN execution can support dynamic 
models, they do with high overheads due to very conservative assumptions and/or expensive analyses at the runtime. For example, TFLite~\cite{TensorFlow-Lite} and MNN~\cite{Ali-MNN}  perform  re-initialization (equivalent of recompilation) 
when the input shape to the model changes.

This paper presents  the first nuanced approach for optimizing DNN 
inference in the presence of dynamic features. Our approach emphasizes reducing inference latency as well as memory requirements -- the latter being 
quite important on the mobile devices we target.
The foundation of our 
approach is an in-depth study of operators that form the basis for modern DNNs. 
These operators are classified into several groups on the basis of how the output 
shapes relate to input shapes and values. 
Based on such a classification, 
we present a data-flow analysis framework, called Rank\footnote{\revisioncr{Rank denotes the number of dimensions in a tensor.}} and Dimension Propagation (RDP) that 
infers shapes and dimensions of intermediate tensors. RDP analysis considers known 
constants, symbolic constants, and expressions involving these. RDP analysis results 
are then used for enabling a number of optimizations, which includes operator 
fusion and fused code generation, static execution planning, runtime memory allocation, 
and multi-version code generation.  
This work integrates RDP and optimizations enabled by it together and builds a comprehensive framework for optimizing Dynamic DNNs, called ~\projectname.
\projectname is extensively evaluated on 10 cutting-edge DNN models with shape dynamism and/or control-flow dynamism. Specifically, these models include the ones for emerging Artificial General Intelligence (AGI)~\cite{goertzel2014artificial} such as StableDiffusion~\cite{rombach2022high} and SegmentAnything~\cite{kirillov2023segany}. Our evaluation results show that \projectname saves $27\%$ to $88\%$ memory consumption and results in $1.7\times$ to $3.9\times$ execution speedup compared with four state-of-the-art product-level DNN execution frameworks (such as ONNX Runtime~\cite{onnxruntime}, MNN~\cite{Ali-MNN}, TVM~\cite{chen2018tvm} with Nimble extension~\cite{shen2021nimble}, and TensorFlow Lite~\cite{TensorFlow-Lite}) that support dynamic DNNs. %

\begingroup
\setlength{\tabcolsep}{4.5pt}
\begin{table}[t!]
\centering
\caption{{\bf Inference overhead for shape dynamism w/ execution re-initialization.} SL:  shape propagation and layout selection. ST: schedule and tuning. Alloc: memory allocation. Infer: inference time. Experiments are conducted on a Samsung Galaxy S21 w/ MNN~\cite{Ali-MNN}.}
\footnotesize{
\begin{tabular}{l|cccc|cccc}
    \toprule
    \multirow{2}{*}{Model} & \multicolumn{4}{c}{CPU latency (ms)} & \multicolumn{4}{c}{GPU latency (ms)} \\
    ~ & SL  & ST  & Alloc & Infer & SL  & ST  & Alloc & Infer \\
    \hline
    YOLO-V6 ~\cite{li2022yolov6} & 6.9 & 1,155  & 2.2 & 476 & 0.8 & 1,678 & 30,605 & 102 \\
    Conformer ~\cite{gulati2020conformer} & 3.8 & 127  & 7.8 & 926 & 3 & 1,021 & 73,170 & 1,193 \\
    CodeBERT ~\cite{feng2020codebert} & 2.3 & 253  & 2.8 & 370 & 1 & 856 & 4,568 & 498 \\
    \bottomrule
\end{tabular}
}
\label{tab:device_execution_comparison}
\end{table}
\endgroup 

In all, this paper makes the following contributions. 
\noindent 
{\bf DNN Operator Classification.} We classify the operators used for modern DNNs (specifically 
150 operators used in ONNX (Open Neural Network Exchange)) into 4 categories, which are {\em Input Shape Determined Output},
{\em Input Shape Determined Output Shape},
{\em Input Shape \& Value Determined Output Shape},
{\em Execution Determined Output}. We formally define these operators and explain their significance for inferring 
ranks and dimensions for the DNNs where the input can be of different sizes and the 
execution is data dependent.  

\noindent 
{\bf Data-Flow Analysis for Rank and Dimension Propagation.  } 
Building on the operator classification, we have developed a static analysis framework 
for propagating shape and size information through a computational graph. This framework, 
called RDP, 
considers both known and symbolic constants as well as expressions involving 
these values. Though somewhat similar to the well-known constant propagation  
analysis~\cite{callahan1986interprocedural}, our work is different 
in having transfer functions specific to the operator (types), supporting 
both backward and forward analyses, and considering not only known and symbolic 
constants but also expressions involving them.  

\noindent 
{\bf Comprehensive Set of Static and Dynamic Optimizations.} Using results 
from RDP analysis, we enable a series of optimizations. First, we enable code 
fusion, including generating multiple versions when sufficient static information 
is not available. Next, we perform execution planning, using the results of RDP to 
partition the original graph, and further using several heuristics based on RDP output. 
Finally, we enable runtime plan generation for memory allocation and also generate 
multiple versions of optimized implementations for individual operators.

\section{Existing Frameworks and  Limitations}

Existing DNN inference engines on mobile devices use two common 
approaches when handling dynamic DNNs.

\noindent{\bf Static Solutions.} Many existing DNN inference engines  
for mobile platforms 
(specifically,  TFLite~\cite{TensorFlow-Lite} and MNN~\cite{Ali-MNN}) support dynamic features by extending their static model execution.  
For handling dynamic input shapes,  this involves either  execution re-initialization when the  input shape changes or, alternatively, 
conservative (maximum) memory allocation when the  input shapes are unknown. 
To handle dynamic control flow, it typically requires 
 the execution of all possible paths, and  stripping out  invalid results.   Not surprisingly, such simplistic handling of dynamic features incurs 
significant execution  and/or memory overhead.  To further illustrate, 
Table ~\ref{tab:device_execution_comparison} shows a performance study of three models (YOLO-V6~\cite{li2022yolov6}, Conformer~\cite{gulati2020conformer}, and CodeBERT~\cite{feng2020codebert}) that can take input with dynamic shapes. MNN~\cite{Ali-MNN} runs these models on a Samsung Galaxy S21 with execution re-initialization to handle varied input shapes. These results  
show that the re-initialization  
usually takes   even significantly longer time  than the inference itself. 
This approach might be acceptable for cases where the overhead of re-initialization can be amortized over a number 
of inference tasks (e.g., certain video processing scenarios). However, many application 
scenarios (across the image, audio, and language processing) involve continuously changing 
inputs.  An alternative way, as also indicated above,  is to conservatively allocate large memory spaces.  However, it incurs significant memory wastage,  which can limit the ability to execute large 
models or to do so efficiently, especially on mobile (or edge)  devices with limited memory.

\noindent{\bf Runtime  Solutions.}
TVM (with  Nimble extension)~\cite{chen2018tvm, shen2021nimble} improves  on the limitations of static solutions 
by  providing a set
of  optimizations within a   virtual machine. An example of this functionality is  a {\em shape function} to infer the output tensor shape and use 
this information  for dynamic memory allocation. 
However, such functions and the subsequent dynamic memory allocation  introduces 
significant execution overhead.

\begin{figure*}[t]
    \centering
    \includegraphics[width=0.9\textwidth]{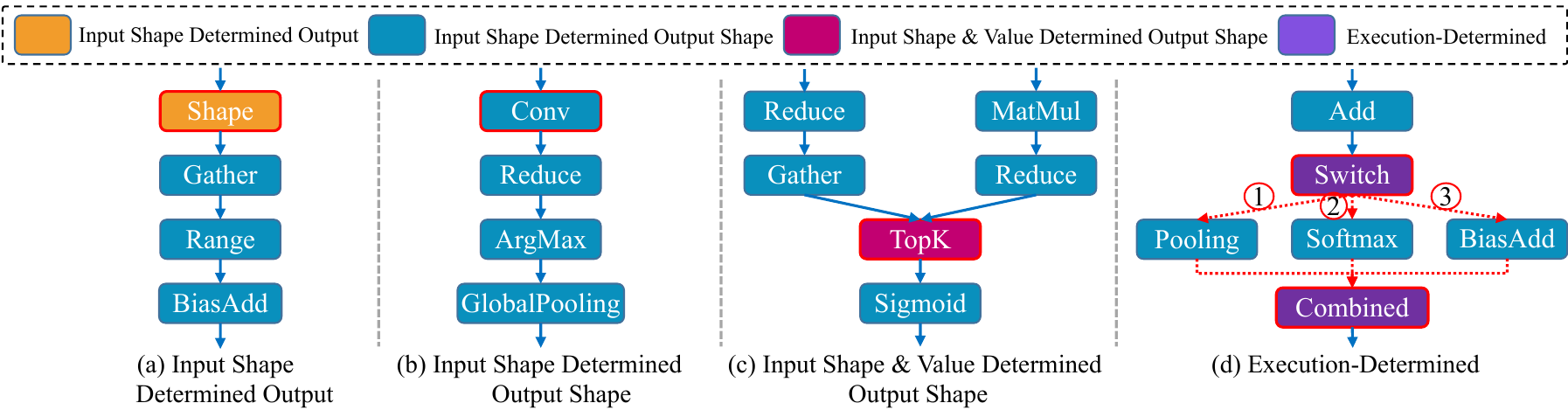}
    \caption{{\bf Different degrees of dynamism.} Each node is a DNN operator. Yellow, blue, red, and purple mean {\em Input Shape Determined Output}, {\em Input Shape Determined Output Shape}, {\em Input Shape \& Value Determined Output Shape}, and {\em Execution Determined Output}, respectively. In (d), {\tt Switch}'s execution path is decided dynamically during runtime and red dot edges represent both the computation dependency and control flow.}
    \vspace{-1em}
    \label{fig:dynamism_examples}
\end{figure*}

\section{Operator Classification based on Dynamism}

Our observation  is that DNN operators have   different dynamism degrees, 
leading to distinct  levels of challenges and opportunities in  optimizing 
them. More specifically,  this work categorizes DNN operators into four types:  
{\em Input Shape Determined Output}, {\em Input Shape Determined Output Shape}, {\em Input Shape \& Value Determined Output Shape}, and {\em Execution Determined Output}. 
This section gives a formal definition.

\paragraph{Background  and Notation.}  
It is common to represent a DNN as 
a Computational Graph, which happens to be a  Directed Acyclic Graph (DAG).   
Each  tensor (which can be an input and/or output)  can be categorized by a 
shape (including dimensions) and the contents or values. 
Each  
operator is denoted as $L^l$, where $l$ is the operator index. Assume $L^l$ has $m$ input tensors (of which, $[1, k)$ are constant tensors while $[k, m]$ are output tensors from previous operators) %
and $n$ output tensors. 
The shape of the input tensor $i$ for the $l^{th}$  operator 
is  denoted as $IS^l_i$ and the corresponding tensor  value can be denoted as  $IV^l_i$,  Similarly, each output tensor's shape and value are  $OS^l_i$ and $OV^l_i$, 
respectively.  
Now, intuitively, a class of functions relates the output shapes and values 
to the input shapes and values --  $F^{fs}$ for the shapes and  $F^v$  for the 
values.

\begingroup
\setlength{\tabcolsep}{6pt}
\begin{table*}[t]
\centering
\caption{{\bf Classification of DNN operators based on dynamism degrees.} Operators are from ONNX (Open Neural Network Exchange)~\cite{onnx-cite}.}
\label{tab:layer_type_classification}
\small
\begin{tabular}{l|l|c}
     \toprule
     {Operator type} & Operators & Representative \\ \hline
     {Input Shape Determined Output}  & \makecell[l]{\opset Shape, ConstantOfShape, Eyelike}  & {\opset Shape} \\ \hline
     \makecell[l]{Input Shape Determined Output Shape}  &  \makecell[l]{\opset Add, AveragePool, Cast, Concat, Conv, Elementwise w/ broadcast, \\ \opset Gather, MatMul, MaxPool, Reduce, Relu, Round, Sigmoid, Softmax}  & {\opset Conv, MatMul} \\ \hline
     \makecell[l]{Input Shape \& Value Determined Output Shape}   & \makecell[l]{\opset Expand, GroupNormalization, MaxUnpool, Onehot, Range, Reshape, \\ \opset Resize, Slice, TopK, Upsample}  & {\opset Reshape, Range} \\ \hline
     {Execution Determined Output}   & \makecell[l]{\opset If, Loop, NMS, Nonzero, <Switch, Combine>$^\dag$}  & {\opset If, Loop} \\ \bottomrule
     \multicolumn{3}{l}{\makecell[l]{{\scriptsize $^\dag$ {\tt <Switch, Combine>} is a pair of customized operators for dynamic control flow that is not defined in ONNX.}}} \\
     
\end{tabular}
\end{table*}
\endgroup

\begin{itemize}[leftmargin=*,noitemsep,nolistsep]
    \item {\bf Input Shape Determined Output:} 
    The output (tensor), which is characterized by both its shape and 
    value, has the following dependence on the input. The output tensor shapes are dependent on the input tensor shapes, whereas  the output tensor values are determined by the input tensor shapes and possibly 
    some of the constant  tensors -- input values do not impact the output.   Examples include  {\opset Shape} and {\opset   EyeLike}. Formally, 
    there is a pair of functions $(F^{fs}, F^v)$, such that:
        \vspace{-0.5em}
        {\small $$OS^l_i \xleftarrow{(F^{fs})} (IS^l_1, \dots, IS^l_m)$$}
        \vspace{-1.6em}
        {\small $$OV^l_i \xleftarrow{(F^v)} (IS^l_1, \dots, IS^l_m), (IV^l_1, \dots, IV^l_{k - 1})$$}
     where $1 \leq k \leq m $. 
    \item {\bf Input  Shape Determined Output Shape: }  Similar to  the 
    previous category, the output shapes depend on the input shapes. However, what 
    is different is that the output values rely  on all the input values (including intermediate and constant input values). Examples 
    include {\opset Conv}, {\opset Add}, and {\opset  Pooling}. The significance of this category, 
    as compared to the next set of categories, is that if the input shape of this operator is known,  compiler optimizations (e.g., operator fusion, execution/memory optimizations) are enabled.  Formally,   
    there is a pair of functions $(F^{fs}, F^v)$, such that:
        \vspace{-0.5em}
        {\small $$OS^l_i \xleftarrow{(F^{fs})} (IS^l_1, \dots, IS^l_m)$$}
        \vspace{-1.2em}
        {\small $$OV^l_i \xleftarrow{(F^v)} (IS^l_1, \dots, IS^l_m), (IV^l_1, \dots,  IV^l_m).$$}
   
    \item {\bf Input Shape \& Value Determined Output Shape:}
    Similar to the previous category, the output values rely on the input shapes and all the input values. The difference is that the output shapes also rely on partial set of input values.
    Examples 
    include {\opset Extend} and  {\opset  Range}). Formally,  there is a pair of functions $(F^{fs}, F^v)$ and a subset of input tensors ($p, \dots, q$) whose values specify the output shape, such that:
        \vspace{-0.5em}
        {\small $$OS^l_i \xleftarrow{(F^{fs})} (IS^l_1, \dots, IS^l_m), (IV^l_p, \dots,  IV^l_q)$$}
        \vspace{-1.5em}
        {\small $$OV^l_i \xleftarrow{(F^v)} (IS^l_1, \dots, IS^l_m), (IV^l_1, \dots,  IV^l_m)$$}
    , where $\ 1 \leq p \leq q \leq m$. If $p \leq q \leq k$, which is identical to {\em Input Shape Determined Output Shape}, and all the dependent input tensors are constant. In such cases, the input shapes can be calculated without knowing other intermediate input tensors. If only the input shape of this operator is known, only partial compiler optimizations with conservative analysis can be applied to it, and full optimizations need  dynamic execution results. 
    \item {\bf Execution Determined Output:}
    Similar to the previous two categories, the output values rely on the input shapes and all the input values.
    Examples include  {\opset Nonzero}  and {\opset  If}.  
    Formally, there is a function $F^v$, such that:
            \vspace{-0.5em}
            {\small $$OV^l_i \xleftarrow{(F^v)} (IS^l_1, \dots, IS^l_m), (IV^l_1, \dots,  IV^l_m)$$}, and the shape of $i$-th output tensor can only be measured after materializing its value:
            \vspace{-0.5em}
            {\small $$OS^l_i \xleftarrow{} SHAPE\_OF(OV^l_i)$$}
    , which means it is not able to know the output shapes until materializing the output tensors (i.e., after executing the layer). Only partial optimization with conservative analysis can be applied to this operator, and full optimizations need  dynamic execution results. 
\end{itemize}

Although these operator types are defined according to {\em forward transfer }, i.e. an output tensor shape and value are related to the input tensor shape and/or value. In practice,  {\em Backward transfer} is also used, i.e., 
we can (and need to) backward propagate the known output shapes (either rank or dimension or both) to the unknown input shapes. For instance, if we know the output shape of {\opset Add}, its input dimension might be 1 or identical to the corresponding output dimension due to broadcasting rules~\cite{broadcasting}.  We define backward transfer functions as:
{\small $$ IS^l_i \xleftarrow{(F^{bs})} (OS^l_1, \dots, OS^l_n). $$}

Table ~\ref{tab:layer_type_classification} shows typical operators in ONNX~\cite{onnx-cite} categorized by the above classification.  
As further illustration, 
Figure~\ref{fig:dynamism_examples} shows four sub-graphs that 
represent operators with different dynamism degrees (marked with red boundary) and their connections.  Figure \ref{fig:dynamism_examples} (a) shows an {\em Input Shape Determined Output} operator {\tt Shape}. Once its input shape is known, its value result can be directly inferred (and in fact, this value can be propagated from {\tt Shape} to {\tt BiasAdd} because all following operators belong to the {\em Input Shape Determined Output Shape} group).
Similarly, Figure ~\ref{fig:dynamism_examples} (b)  implies  that if the input shape to {\tt Conv} is known, this shape information could be propagated to the entire sub-graph because all operators in this sub-graph belong to the {\em Input Shape Determined Output Shape} group. For the cases represented 
in both (a) and (b), even if the exact shape is unknown, it is still possible for us to perform compiler optimizations such as operator fusion and fused code generation, execution order optimization, and memory optimization, which will be elaborated in the next Section.   
In Figure~\ref{fig:dynamism_examples} (c), the output shape of {\tt TopK} depends on its input value (which is the left predecessor's branch in the example), i.e., the output shape of {\tt TopK} (and its successors) is unknown until its left predecessor branch is executed. 
Figure~\ref{fig:dynamism_examples} (d) represents a sub-graph involving a dynamic control flow. {\tt Switch} results decide if path \ding{172}, \ding{173}, and/or \ding{174} will be taken, and  {\tt Combine} merges the results from executed paths. Both (c) and (d) require dynamic execution, thus is  more difficult to  optimize statically. 

\noindent 
{\em Discussion.} Although the examples in Table~\ref{tab:layer_type_classification} and Figure~\ref{fig:dynamism_examples}  mentioned above simply classify each operator into one category, there 
are additional considerations.  For example, an {\tt Upsample} operator may belong to either {\em Input Shape Determined Output Shape} or {\em Input Shape \& Value Determined Output Shape} depending on whether some of the input tensors are constant or not. 
Therefore, with  constant propagation, an operator may transform from a more dynamic classification to a less dynamic one, offering us more aggressive optimization opportunities. This has motivated
certain aspects of  \projectname.
\noindent\revisioncr{
In addition, because our operator classification essentially models the dynamism degree of an operator by studying its computation logic and input/output tensor shapes and values, it is possible to create an automatic and generic tool based on existing intermediate representations like tensor expression (e.g., TVM expression) to categorize operators into different classifications.
}

\section{Design of \projectname}

Based on the DNN operator classification introduced above, 
\projectname introduces a new static data-flow analysis framework to infer the intermediate result tensor shape. Such an analysis is the enabler of several optimizations, which are dynamic DNN operator fusion, execution path planning, memory planning, and multi-version code generation. 
\revisioncr{
All of these optimizations ensure a deterministic running sequence and a consistent output, given a particular input.
}
At a high level, our approach does not require conservative static assumptions or runtime overheads, thus providing significant improvement over the existing state-of-the-art.

\subsection{Pre-Deployment  Data-Flow Analysis}

\begin{figure}[t]
    \centering
    \vspace{0.5em}
    \includegraphics[width=0.45\textwidth]{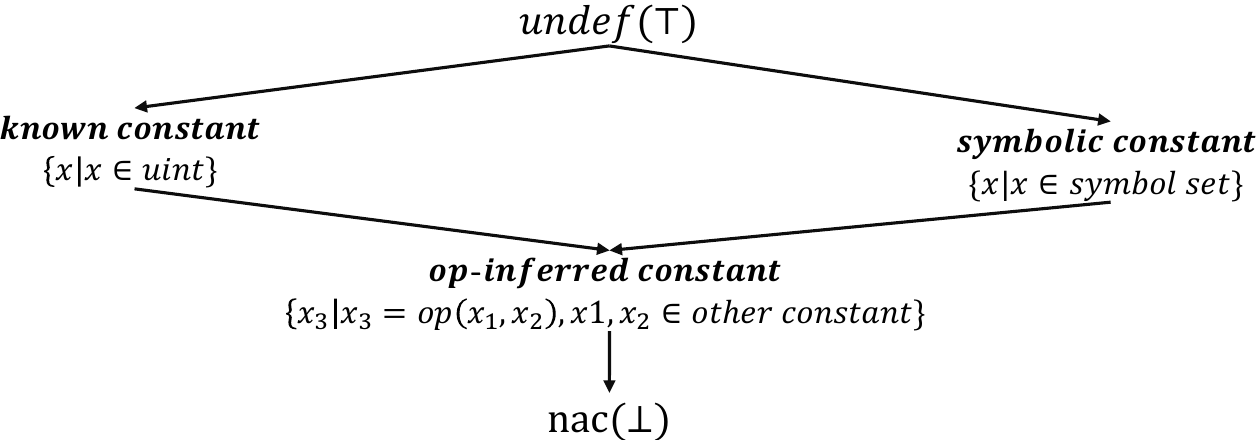}
    \caption{{\bf Domain of RDP dataflow analysis.} It includes known, symbolic, and operation-inferred constants that form a lattice. }
    \label{fig:lattice_def}
    \vspace{-1em}
\end{figure}

\begin{figure*}[t]
    \centering
    \vspace{0.25em}
    \includegraphics[width=0.9\textwidth]{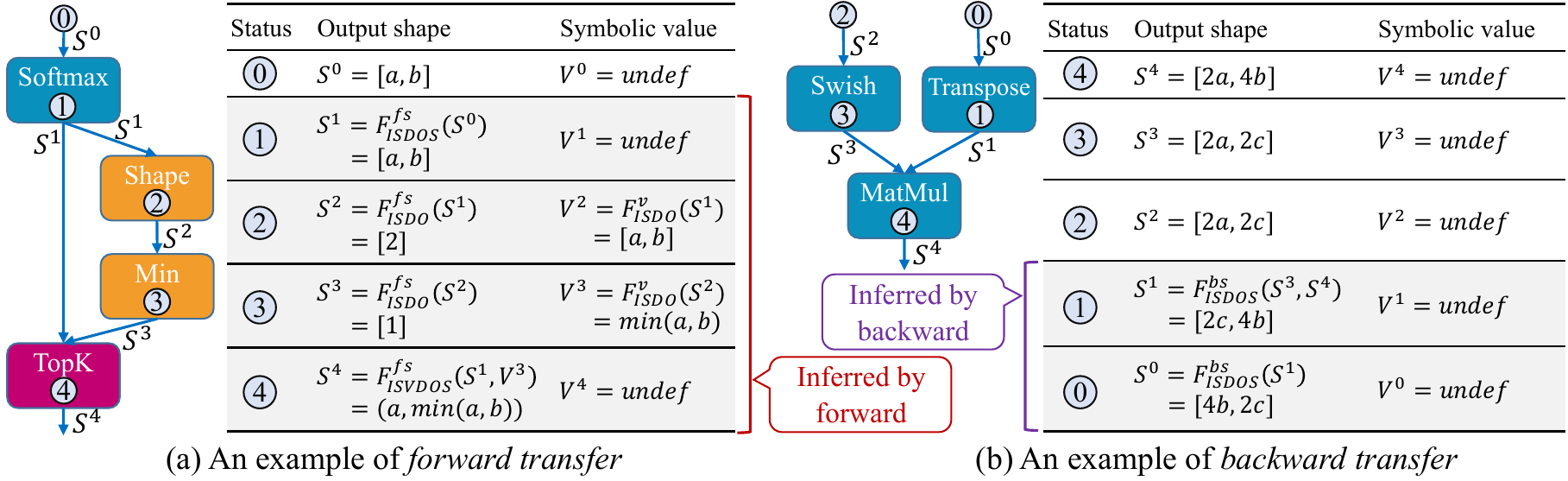}
    \caption{{\bf Examples of forward and backward transfer.} Each node is an operator. Yellow, blue, and red mean {\em Input Shape Determined Output}, {\em Input Shape Determined Output Shape}, and {\em Input Shape \& Value Determined Output Shape}, respectively. Ids (e.g., \ding{172}) indicate the location where transfer functions apply and their applying orders for a forward transfer (a backward transfer reverses this order). S and V equations map values in the RDP domain to the shape and value of each tensor, in which, F denotes the transfer function. {\em fs} and {\em bs} of F denote forward and backward, and F's subscript is a short form of its type (e.g., {\em ISDOS} means {\em Input Shape Determined Output Shape}). 
    }
    \vspace{-0.25em}
    \label{fig:forward_backward_example}
\end{figure*}

To facilitate static optimizations for dynamic DNNs, 
a critical requirement is  knowing (possibly symbolically) 
the intermediate result tensor shape (i.e., rank and dimension). 
Our key observation is that {\em for many operators and operator combinations  (e.g., an Input Shape Determined Output operator and an Input Shape Determined Output Shape operator), even without knowing the input tensor shape, it is still possible to infer the shape of the intermediate result tensor to a certain degree.}  
Our framework is based on this observation and  is  called operator Rank and Dimension Propagation, or 
RDP.  While RDP has certain similarities with the classical (symbolic) constant  propagation   
frameworks~\cite{callahan1986interprocedural}, it needs to deal with nuances of the DNN 
operations and the computational graph. 
  RDP  also considers operations over multiple (symbolic) constants as a possibility in its lattice 
  and requires iterative forward and backward analysis. 

\paragraph{Formal Definition of Operator Rank and Dimension Propagation (RDP).}
The entire RDP algorithm is expressed as a four-tuple $<G,D,L^{\prime},F >$. 

\begin{itemize}
    \item $G$ is an {\em extended computational graph} (a DAG),  with control-flow  operators 
    <Switch, Combine>). If this extended computational graph involves multiple branches that all need to be executed, we assume the execution order is always from left to right. It is easy to prove that $G$  is equivalent to a control-flow graph on operators, which serves as the foundation of this data-flow analysis. 
    \item $D$ is the direction of the data flow, which can be  FORWARD and BACKWARD.   
    Unlike most classic data-flow  formulations,  e.g., constant propagation or reaching definitions, RDP  iteratively processes $G$ in  forward and backward directions  until the 
    results converge.  This is because the shape of a tensor could be inferred from its producing operator and/or consuming operator, and their inference results should be the same to guarantee the correctness of this DNN execution.
    
    \item $L^{\prime}$ itself  is a three-tuple $<V',\wedge,m >$. $ V^{\prime}$ 
     is the domain of values (also shown in Figure~\ref{fig:lattice_def}) and includes 
    known constants, symbolic constants, and operation-inferred constants that form a  lattice. 
    The lattice also includes  undefined ($undef$)  as the top ($\top$) of the 
    lattice and Bottom ($\bot$)  which is  not-a-constant ($nac$). %
    $\wedge$ is a meet operator, which follows the common definition  for product lattice. $m$ is a map function mapping values in lattice to two variables, Shape (S) and Value (V),  representing these for the intermediate tensor. 
    \revisioncr{More specifically, RDP is a type of data-flow analysis, where L$^{\prime}$ describes its analysis scope, i.e., how to map each shape and value property to a kind of constant that forms a lattice structure (in Figure~\ref{fig:lattice_def}). This lattice guarantees that the analyzed properties of RDP follow lattice theory, so RDP analysis will converge with a unique solution.}
    \item $F: V' \rightarrow V'$ is the domain for transfer functions.  $F$ is designed for each operator (type) and transfers  the Shape (S) and Value (V) from the input tensor to the output tensor based on the operator type. 
    Similar to data-flow analysis 
    for {\em constant propagation} RDP has two kinds of transfer functions, {\tt Update}, and {\tt Merge}. {\tt Update}  transfers from the input tensor to the output tensor for an individual operator; while {\tt Merge} operates on branch control flow and merges (output) tensors from multiple possible execution paths. Because RDP has both FORWARD and BACKWARD directions, $F$ also contains transfer functions for both directions. 
\end{itemize}

\begingroup
\setlength{\tabcolsep}{2pt}
\begin{table}[t]
\centering
\caption{\revisioncr{Illustration of forward and backward transfer functions for different operator types.}}
\footnotesize{
\revisioncr{
\begin{tabular}{l|cc|cc}
    \toprule
    \multirow{2}{*}{Type} & \multicolumn{2}{c|}{Forward} & \multicolumn{2}{c}{Backward} \\
    ~                     & Shape & Value                & Shape & Value                \\
    \hline
    Input Shape Determined Output   & $F^{fs}_{ISDO}$   & $F^{fv}_{ISDO}$   & $F^{bs}_{ISDO}$   & $F^{bv}_{ISDO}$ \\
    Input Shape Determined Output Shape  & $F^{fs}_{ISDOS}$  & $F^{fv}_{ISDOS}$  & $F^{bs}_{ISDOS}$  & $F^{bv}_{ISDOS}$ \\
    Input Shape \& Value Determined Output Shape & $F^{fs}_{ISVDOS}$ & $F^{fv}_{ISVDOS}$ & $F^{bs}_{ISVDOS}$ & $F^{bv}_{ISVDOS}$ \\
    Execution Determined Output    & $F^{fs}_{EDO}$    & $F^{fv}_{EDO}$    & $F^{bs}_{EDO}$    & $F^{bv}_{EDO}$ \\
    \bottomrule
\end{tabular}
}
}
\label{tab:transfer_function}
\end{table}
\endgroup

\noindent{\em Transfer Function Examples.}
\revisioncr{
\projectname contains 16 types of {\tt Update} transfer functions. These functions are based on the classification of the operator's dynamism degree, as detailed in Table ~\ref{tab:transfer_function}. The table includes four dynamic degrees, covering two directions (forward and backward) and two types of propagation (shape and value). 
During forward transfer, each operator employs its shape and value transfer functions in accordance with its associated dynamism degree to infer the shape (e.g., $F^{fs}_{ISDO}$) and value (e.g., $F^{fv}_{ISDO}$) of its output tensor, respectively. 
Similarly, $F^{bs}_{ISDO}$ and $F^{bv}_{ISDO}$ serve as two examples of a backward transfer function for shape and value during the backward transfer process.
}
Figure~\ref{fig:forward_backward_example} illustrates several common ones. The left-hand side (Figure~\ref{fig:forward_backward_example} (a)) shows an example with four forward transfers that employ three types of {\tt Update} transfer functions. 
Similarly, the right-hand side (Figure~\ref{fig:forward_backward_example} (b)) shows an example with two backward transfer functions that belong to the same type. 
A point worth noting is that the appropriate transfer function to apply to an operator depends not only on the computational graph but also on the constants inferred during the RDP analysis process, which determines the dynamism classification of the operator.
The  {\tt Merge} transfer function is straightforward -- it merges the S-map and V-map from multiple control-flow branches based on the lattice in Figure~\ref{fig:lattice_def}. 
Table~\ref{tab:propagation_definition} summarizes the key components of RDP.

\begingroup
\setlength{\tabcolsep}{5pt}
\begin{table}[t]
\centering
\caption{{\bf Definition of Rank and Dimensions Propagation (RDP).}}
\footnotesize{
\begin{tabular}{c|c||c|c}
    \toprule
    Notation       & Definition & Notation       & Definition \\
    \hline
    Domain & Tensor Rank and Dimensions & Direction & {\em Forward}, {\em Backward}\\
    Forward & $OUT(L) = F^{fs}_{P \in pred(L)}(P)$ & Backward & $IN(L) := F^{bs}(OUT(L))$  \\
    Initial & $OUT[L] = undef$ & Terminate & No more changes \\
    \bottomrule
\end{tabular}
}
\label{tab:propagation_definition}
\end{table}
\endgroup

\begin{algorithm}[t]
\DontPrintSemicolon
\footnotesize
\SetNoFillComment
\caption{RDP's Optimized Chaos Algorithm}\label{alg:dfa_analysis}
\SetKwInput{ANALYZE}{Func rdp\_analysis}
\SetKwInput{FORWARD}{Func forward\_transfer}
\SetKwInput{BACKWARD}{Func backward\_transfer}
\SetKw{Return}{return}
\SetKwRepeat{Do}{do}{while}%
\ForEach{node in ecg.sorted\_node}{
  mark\_as\_undef(node) \tcc{Initialize as $undef$}
}
set\_model\_input\_shape(ecg)\;

\Do{changed}{
  changed $\gets$ false \;
  \tcc{Traverse the Depth-first sorted nodes}
  \ForEach{node in ecg.sorted\_node}{    
    predecessors $\gets$ predecessor\_of(node) \;      
    successors $\gets$ successor\_of(node)     \;
    \If{node.type is Combine}{ 
      \tcc{Merge the rank and dims for Combine}
      changed $\mid=$ node.merge(predecessors) \;
    }
    \If{node.type is Switch or Combine}{
      continue \tcc{Transit to all successors}
    }
    \tcc{1 Forward transfer to current node}
    changed $\mid=$ forward\_transfer(node, predecessors) \;
    \tcc{2 Backward transfer to predecessors}
    \ForEach{pred in predecessors}{
      changed $\mid=$ backward\_transfer(node, pred) \;
    }
    \tcc{3 Update for Input Shape Determined Output}
    \If{node.type $\in$ Input Shape Determined Output}{      
      \If{node.shape $\notin (undef, nac)$}{
        node.value $\gets$ get\_symbolic\_value(node.shape)
      }        
    }
  }
}

\vspace{0.15cm}

\FORWARD{node, preds}
\If{all(node.outputs.shape $\notin$ undef)}{  
  \Return False \tcc{Outputs are not in $undef$}
}
pred\_shapes, pred\_values $\gets$ shape\_of(preds), value\_of(preds)
\Switch{node.op\_type} {
  \Case{`Input Shape Determined Output'} {
    \tcc{Only depends on the first input shape}
    \Return FT\_ISDO(node, pred\_values[0])\;
  }
  \Case{`Input Shape Determined Output Shape'} {
    \Return FT\_ISDOS(node, pred\_shapes)\;
  }
  \Case{`Input Shape \& Value Dependent Output Shape'} {
    \Return FT\_ISVDOS(node, pred\_shapes, pred\_values)\;
  }
  \Case{`Execution-Determined'} {
    \tcc{Assign $nac$}
  }
}
\Return False\;

\vspace{0.35cm}

\BACKWARD{node, pred}
\If{all(pred.outputs.shape $\notin$ undef)}{  
  \Return False \tcc{Outputs are not in $undef$}
}
\tcc{Similar to forward\_transfer}
\end{algorithm}

\paragraph{RDP Solution. }  
The method is shown as Alg.~\ref{alg:dfa_analysis} and involves 
 applying the  transfer functions (F) to the extended computational graph (G) along  the 
two directions iteratively.   
Elaborating on Alg.~\ref{alg:dfa_analysis}, it  first sorts the nodes (i.e., operators) in the computational graph $G$  with the dept-first 
order  and initializes the output shape- and value-maps of each node as {\em undef} (Line 1 to Line 2). 
It next processes each node ($n$) by applying forward transfer functions to $n$'s predecessors' output shape- and value-maps (i.e., $n$'s input shape- and value-maps) (Line 13). 
Moreover, it  propagates  $n$'s output shape- and value-maps to $n$'s predecessors' output shape- and value-maps by backward transfer functions if any predecessors have {\em undef} analysis results (Line 14 to Line 15). These forward and backward transfer functions are defined based on the dynamism classification of DNN operators (as shown in Line 20 to Line 32). Alg.~\ref{alg:dfa_analysis} needs to process two specific types of nodes (operators): i) control-flow nodes (like {\tt Combine} or {\tt Switch}), for which, it needs to call the {\tt Merge} function to merge analysis results from multiple control-flow paths %
(Line 9 to Line 10), and ii) Input Shape Determined Output nodes, for which, it assigns a symbolic constant to the value map to facilitate subsequent analysis (Lines 16 to 18).  Alg.~\ref{alg:dfa_analysis} continues processing  nodes in $ G$ until no updates happen on any node's shape-/value-maps.
Similarly to other data-flow analysis, RDP follows {\em Lattice Theory}~\cite{kildall1973unified}, so an optimized chaos implementation  (based on worklist) is guaranteed  to  converge.

\subsection{Operator Fusion for Dynamic DNN based on RDP} 
\label{subsec:fusion}

Though fusion has been a successful optimization on DNNs~\cite{niu2021dnnfusion}, 
it is also known to be very hard to implement on  dynamic DNNs~\cite{shen2021nimble}. A frequent issue is that without knowing the tensor shape of two operators, the DNN compiler either cannot fuse them at all  or has to generate a large number of code versions, each for a possible  combination of  shapes 
for the two operators. In fact, as often more than  two operators are merged, 
the possible combinations for which separate code should be generated increase rapidly. 
Our proposed RDP analysis can address  this issue by using 
(possibly symbolic) shape information. Information such as the two operators having 
tensors of the same shape can enable and/or simplify fusion, even if the exact dimensions 
are not known till runtime.

\begin{figure}[t]
    \centering
    \vspace{0.5em}
    \includegraphics[width=0.45\textwidth]{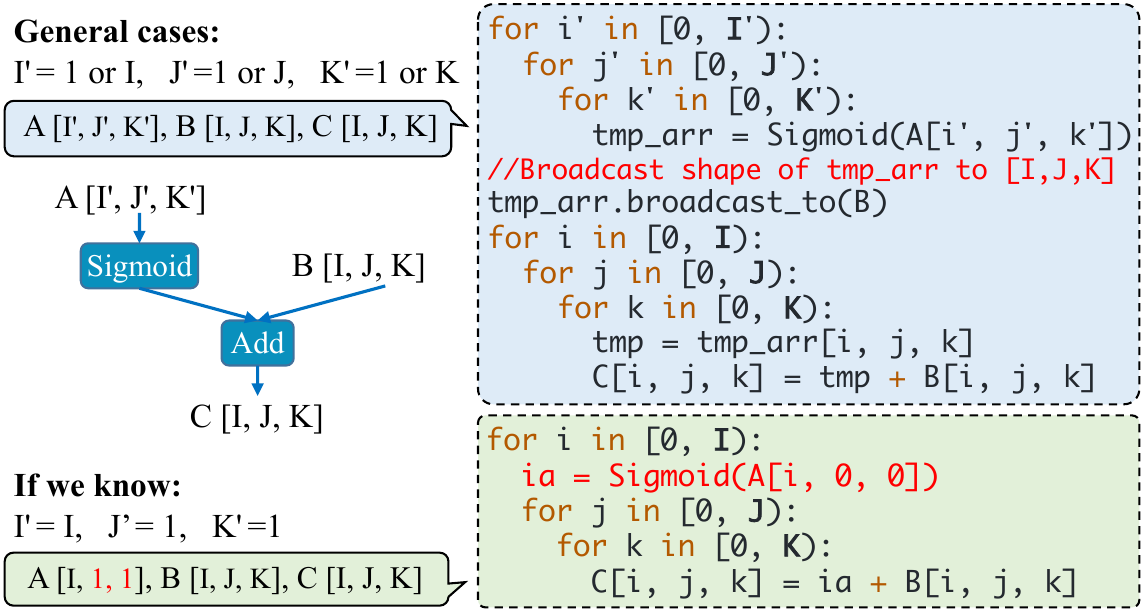}
    \caption{{\bf Operator fusion with dynamic shapes.} The top code snippet shows that fusion is not feasible because of broadcasting~\cite{broadcasting}. Specifically, {\em Add} requires A's indices $I'$, $J'$, and $K'$ to be either 1 or $I$, $J$, and $K$, resulting in 8 fusion scenarios. With RDP, such fusion is feasible (shown in the below code snippet). This fusion 
    significantly reduces intermediate result materialization requirements.}
    \label{fig:fusion_challenge}
    \vspace{-0.5em}
\end{figure}

Figure~\ref{fig:fusion_challenge} shows a simplified example with two common DNN operators ({\tt Sigmod} and {\tt Add}) on  tensors with shapes not known till runtime. 
{\tt Sigmod} takes an input tensor A with a dynamic shape of [I', J', K']. {\tt Add} performs an element-wise addition on {\tt Sigmod}'s output and another input tensor B, whose 
shape happens to be [I, J, K]. 
Now, if A and B are of different shapes, a shape broadcast operation on the output tensor of {\tt Sigmod} needs to be conducted immediately before the element-wise addition. Without our RDP analysis, the dynamic shape of A and B (and the possible shape broadcast operation) prevents the DNN compiler from fusing these two operators in an efficient way, i.e., the compiler either generates code without fusion (as shown in the blue box of Figure~\ref{fig:fusion_challenge}), or generates multiple code versions (8 versions for this example) and selects a version during the runtime. Assuming our RDP analysis result is I' = I, J' = 1, and K' = 1, i.e., a mix of symbolic constant (I) and known constant, the DNN compiler can further generate a unique version of fused code (as shown in the green box of Figure~\ref{fig:fusion_challenge}).
\projectname incorporates RDP and the above operator fusion based on RDP into a state-of-the-art operator fusion for static DNNs (DNNFusion~\cite{niu2021dnnfusion}
) 
to generate the fusion plan and optimized fused code.

\subsection{Static Execution Planning based on RDP}
\label{subsec:sep} 
A computational graph (DAG)  typically allows for several different orderings  for the execution of operators. The choice of ordering has an impact on the peak memory usage (for intermediate 
results), which further has consequences for 
cache performance and the execution latency.  
There has been previous work on this problem, which has in fact shown that 
 generating an {\em optimal}  execution plan (by a metric like memory consumption) 
 is  an NP-complete problem~\cite{ahn2020ordering}.  Thus, 
 choosing an optimal plan can be difficult for modern large DNNs with hundreds or even thousands of operators. 
 
 The dynamic properties (e.g., dynamic shapes and control flow) further complicate this problem.  
 In \projectname, we  develop a series of heuristics driven by the use 
 of proposed RDP analysis. The overall idea is that since a globally optimal solution is almost infeasible, an approach based on {\em graph partitioning} 
 is justified.  It turns out that the results of  RDP are able to guide both  graph partitioning   
 and choice of solution within each sub-graph. 
Particularly, we observe  that known constants, symbolic constants, op-inferred constants, and  
$\bot$ or {\em nac} progressively increase the impediment on the generation of an optimal execution plan.
More specifically, for a sub-graph  $sg$ with a limited number of operators: 

    First, if the shape of all tensors in $ sg$  are known constants, the optimal execution plan  for $sg$ can be obtained statically   by an exhaustive search -- a limited size of $sg$ can 
    further make such a search feasible. 
    Second, if the shape of tensors in $ sg$  are mixed known constants, symbolic constants, and op-inferred constants, it is still possible to  compare the memory requirements and 
    thus generate a (close to) optimal execution plan. This is especially true if these shapes are derived from the same set of symbolic  constants. 
        Third, if an operator has an {\em nac} output tensor shape, it disables further analysis and execution planning. Such operators, it turns out, provide an opportunity to partition the 
    original graph into sub-graphs that can be independently analyzed.

\subsection{Other Optimizations}\label{sec:other-opt}

\subsubsection{ Memory Allocation Plan}\label{sec:memory-alloc-plan}

Besides execution (order) planning, {\em memory planning} of DNNs is also a critical 
step~\cite{ahn2020ordering, pisarchyk2020efficient}. A {\em memory allocation plan}, which decides where in a linear memory space each 
intermediate tensor is allocated, and when it is deallocated, 
can restrict peak memory usage and improve execution performance -- the latter by reducing memory fragmentation, avoiding memory movement, and limiting memory allocation/de-allocation. 
In contrast to execution planning that (even for dynamic DNNs) can be carried out 
at compilation time, memory planning for dynamic DNNs can only be performed at execution time when all tensor sizes are known. Memory planning of static DNN execution has also been proved NP-complete~\cite{ahn2020ordering}, while DNN model dynamism further complicates memory planning. 

Existing memory planning methods  for dynamic DNN execution (e.g., Nimble~\cite{shen2021nimble}) have addressed this. 
Without knowing the exact tensor shapes, the  methods usually rely on a greedy strategy ~\cite{pisarchyk2020efficient}, (e.g., finding the minimal memory slot currently available that can hold the new tensor). 
In comparison, we use RDP results and  the following two key insights. 
First, we base our approach on   sub-graphs generated by our static execution planning method. It turns out that for sub-graphs with known constant shapes, as well as those with symbolic/op-inferred constant shapes that are defined solely by the input tensor of the sub-graph, 
the peak memory requirement  can be inferred from static RDP analysis results and subsequent execution plan generation.  

Second, we have observed that for most sub-graphs, the memory requirement decreases monotonically in both forward and backward directions from the location  
in the graph with peak memory usage. Therefore, initiating memory planning from the peak memory consumption location and traversing in the forward and backward directions, 
and picking  the  available memory slots for reuse works as a good strategy, 
and does not lead to the need for  extra memory space.

Based on these insights, a lightweight greedy approach that starts from the peak memory requirement location can help to find optimal memory usage for many/most sub-graphs.
Our  evaluation   (details 
omitted because of space limits) on ConvNet-AIG~\cite{Veit2018} shows that our RDP-based memory allocation plan requires $1.05\times$ of optimal peak memory consumption (that results from an exhaustive search); while the one based on the greedy strategy mentioned above (MNN) requires $1.16\times$ of optimal peak memory consumption.

\subsubsection{RDP-based Multi-Version Code Generation}\label{sec:multi-verion-code-gen} 

As we discussed in Section~\ref{subsec:fusion}, RDP analysis enables and/or simplifies operator 
fusion by revealing (possibly symbolically) tensor shapes. In cases where a single 
(fused) version is not feasible, one  of the advantages of the information 
obtained through RDP is that the number of different versions of the fused code generated can be reduced significantly.

\projectname further benefits from this property of RDP by generating {\em multi-version code}  to optimize {\em hotspot operators}  (e.g., CONV and GEMM) that dominate the DNN execution. Prior efforts~\cite{ TensorFlow-Lite,Ali-MNN}  have shown that the optimization opportunities for these operators depend on the shapes and sizes of the input/output tensors. Therefore, for static DNN executions, existing frameworks (such as TensorFlow Lite~\cite{TensorFlow-Lite} and MNN~\cite{Ali-MNN}) usually employ multi-version codes that involve different  optimizations (e.g., tiling, unrolling, choice of the number of thread blocks, etc.). 
However, this optimization is challenging for dynamic DNNs because an unknown tensor shape and/or tensor size implies that too many versions will be needed. 
The tensor shape (or shape relations)  provided by RDP help to generate code for more specific tensor shapes only, thus resulting in fewer code versions.

\begingroup
\setlength{\tabcolsep}{6.5pt}
\renewcommand\arraystretch{1.0}
\begin{table*}[t!]
\centering
\caption{{\bf Memory consumption (allocated for intermediate results)  for ONNX Runtime, MNN, TVM with Nimble extension (TVM-N), and \projectname on a mobile CPU.}  ``-'' means this model is not supported by the framework yet. ``S'' stands for shape dynamism, and ``C'' represents for control-flow dynamism.
}
\label{tab:eva_memory_report}
\small
\begin{tabular}{lcc|c|c|cc|cc|cc|cc}
     \toprule
     \multirow{2}{*}{Model} & \multirow{2}{*}{\#Layers} & \multirow{1}{*}{Model} & \multirow{2}{*}{Dynamism} & \multirow{2}{*}{Input Type} & \multicolumn{2}{c|}{ORT (MB)} & \multicolumn{2}{c|}{MNN (MB)} & \multicolumn{2}{c|}{TVM-N(MB)} & \multicolumn{2}{c}{{\bf \projectname (MB)}} \\ \cline{6-13}
     ~ & ~ & Size (MB) & ~ & ~ & Min & Max & Min & Max & Min & Max & Min & Max \\ \hline     
     StableDiffusion~\cite{rombach2022high} & 407 & 137 & S  & Text + Image & 186 & 342 & 124 & 376  & - & - & 92 & 271 \\
     SegmentAnything~\cite{kirillov2023segany} & 857 & 17 & S  & Text + Image & -  & - & - & -  & - & - & 16 & 22 \\
     Conformer~\cite{gulati2020conformer} & 1,703 & 303 & S  & Audio & -  & - & 61 & 78  & - & - & 49 & 58 \\
     CodeBERT~\cite{feng2020codebert} & 985 & 502 & S  & Text & 32  & 75 & 25 & 54  & - & - & 21 & 41 \\
     YOLO-V6~\cite{li2022yolov6} & 599 & 239 & S  & Image & 288 & 430 & 148 & 404  & 964 & 1,103 & 89 & 206  \\
     SkipNet~\cite{wang2018skipnet} & 549 & 103 & S  + C & Image  & 168 & 597 & 27 & 124  & 522 & 700 & 18 & 86 \\
     DGNet~\cite{li2021dynamic} & 847 & 91 & C & Image & 37  & 37 & 76 & 76  & - & - & 23 & 29 \\
     ConvNet-AIG~\cite{Veit2018} & 282 & 104 & S  + C & Image & 168 & 423 & 33 & 109  & 557 & 646 & 26 & 77  \\
     RaNet~\cite{yang2020resolution} & 2,617 & 525 & S  + C & Image & 675  & 1275 & 166 & 675  & - & - & 86 & 452 \\
     BlockDrop~\cite{wu2018blockdrop} & 439 & 179 & S  + C & Image & 242 & 460 & 35 & 105  & 523 & 723 & 24 & 69 \\ \hline
     \multicolumn{5}{c|}{\bf Geo-mean memory consumption normalized by \projectname$^\star$} & \multicolumn{2}{c|}{\bf 3.64$\times$} & \multicolumn{2}{c|}{\bf 1.37$\times$} & \multicolumn{2}{c|}{\bf 8.62$\times$} & \multicolumn{2}{c}{1} \\
     \bottomrule
     \multicolumn{13}{l}{\makecell[l]{\scriptsize $^\star$ This normalized geo-mean memory consumption is calculated by 1) averaging the memory usage of runs with all input samples for each model, 2) calculating the geo-mean \\ \scriptsize of the average memory usage of all models, and 3) normalizing with \projectname's geo-mean memory usage.}}\\
     \vspace{-2em}
\end{tabular}
\end{table*}
\endgroup

\revisioncr{
More specifically, \projectname relies on an auto-tuner based on Genetic Algorithm to generate the exploration space (e.g., tiling shapes, loop permutation, and unrolling settings) for kernel code generation as DNNFusion~\cite{niu2021dnnfusion}. One feature of this auto-tuner is the more effective exploitation of parallelism available in the hardware. To tackle the challenge of dynamic shapes, \projectname employs a multi-version approach, where the versions are chosen based on empirical evidence relating to the impact of different shapes on performance. For instance, our auto-tuner considers fat, regular, and skinny matrices for both GEMM and CONV kernels.
}

\begingroup
\setlength{\tabcolsep}{5.3pt}
\renewcommand\arraystretch{1.0}
\begin{table*}[t!]
\centering
\caption{{\bf End-to-end execution latency comparison among ONNX Runtime, MNN, TVM-N, and \projectname on mobile CPU and mobile GPU.} ``-'' means this model is not supported by the framework yet. }
\label{tab:eva_performance_report}
\small
\begin{tabular}{l|cc|cc|cc|cc|cc|cc|cc|cc}
     \toprule
     \multirow{3}{*}{Model} & \multicolumn{4}{c|}{ORT (ms)} & \multicolumn{4}{c|}{MNN (ms)} & \multicolumn{4}{c|}{TVM-N (ms)} & \multicolumn{4}{c}{{\bf \projectname (ms)}} \\ \cline{2-17}
     ~ & \multicolumn{2}{c|}{CPU} & \multicolumn{2}{c|}{GPU} & \multicolumn{2}{c|}{CPU} & \multicolumn{2}{c|}{GPU} & \multicolumn{2}{c|}{CPU} & \multicolumn{2}{c|}{GPU} & \multicolumn{2}{c|}{CPU} & \multicolumn{2}{c}{GPU} \\
     ~ & Min & Max & Min & Max & Min & Max & Min & Max & Min & Max & Min & Max & Min & Max & Min & Max \\ \hline     
     StableDiffusion~\cite{rombach2022high} & 179 & 2,115 & 217 & 2,076 & 189 & 1,287 & 159 & 1,252 & - & - & - & - & 152 & 733 & 105 & 530 \\
     SegmentAnything~\cite{kirillov2023segany} & - & - & - & - &  - & - & - & - & - & - & - & - &  66 & 108 & 42 & 71 \\
     Conformer~\cite{gulati2020conformer} & - & - & - & - & 51 & 300 & 265 & 498 & - & - & - & - & 40 & 225 & 35 & 150 \\
     CodeBERT~\cite{feng2020codebert} & 141 & 752 & - & - & 125 & 1,265 & - & - & - & - & - & - &  102 & 452 & 72 & 366 \\
     YOLO-V6~\cite{li2022yolov6} & 174 & 1,386 & 155 & 733 & 168 & 925  &  47 & 287 & 251 & 2,108 & - & - & 118 & 546 & 33 & 178 \\
     SkipNet~\cite{wang2018skipnet}  & 111 & 841 & - & - & 92 & 789 & 116 & 363 & 109 & 974 & - & - &  41 & 633 & 29 & 253 \\
     DGNet~\cite{li2021dynamic}  & 122 & 122 & 127 & 127 & 67 & 67 & 211 & 211 & - & - & - & - & 32 & 56 & 23 & 42 \\
     ConvNet-AIG~\cite{Veit2018} & 90 & 693 & - & - & 88 & 731 & 96 & 305  & 98 & 947 & - & - &  46 & 526 & 22 & 203 \\
     RaNet~\cite{yang2020resolution} & 102 & 641 & - & - & 139 & 663 & 114 & 208  & - & - & - & - & 63 & 403 & 31 & 150 \\
     BlockDrop~\cite{wu2018blockdrop} & 153 & 1,199 & - & - & 145 & 1,421 & 139 & 468 & 185 & 1,622 & - & - & 79 & 668 & 42 & 295  \\ \hline
     \multicolumn{1}{l|}{\bf Geo-mean latency$^\star$} & \multicolumn{2}{c|}{\bf 2.5$\times$} & \multicolumn{2}{c|}{\bf 3.9$\times$} & \multicolumn{2}{c|}{\bf 1.7$\times$} & \multicolumn{2}{c|}{\bf 2.3$\times$} & \multicolumn{2}{c|}{\bf 2.7$\times$} & \multicolumn{2}{c|}{-} & \multicolumn{2}{c|}{1} & \multicolumn{2}{c}{1} \\
     \bottomrule
     \multicolumn{17}{l}{\makecell[l]{\scriptsize  $^\star$ This normalized geo-mean execution latency is calculated by 1) averaging the execution latency of runs with all input samples for each model, 2) calculating the geo-mean \\ \scriptsize of the average execution latency of all models, and 3) normalizing with \projectname's geo-mean execution latency.}} \\
\end{tabular}
\end{table*}
\endgroup

\section{Evaluation}

\projectname is implemented by extending an existing DNN execution framework (DNNFusion~\cite{niu2021dnnfusion})
that supports static DNN execution only.
This section evaluates the performance of \projectname by comparing it with four  state-of-the-art frameworks. These frameworks are  ONNX Runtime (ORT) ~\cite{onnxruntime} (V1.14.1), 
MNN~\cite{Ali-MNN} (Vdcb080c), TVM~\cite{chen2018tvm} w/ Nimble extension (TVM-N)~\cite{shen2021nimble} (V7831a79), and TFLite~\cite{TensorFlow-Lite} (V2.11.1). 
ORT, MNN, and TVM-N support shape dynamism, while for DNNs with control flow, they execute all possible branches and strip out invalid ones. 
For fairness, this section also shows a performance comparison between \projectname and MNN by disabling \projectname's {\em <Combine, Switch>} control-flow support and adopting the same ``execute-all, strip-out-invalid'' strategy. 
TFLite supports dynamic input shapes with memory re-initialization; 
however, it cannot run most of our dynamic models properly because it usually fails on some input shapes. It does not support
dynamic control flow either as required by most of the  models we target.
Thus, we use TFLite  as a baseline for comparing DNN executions with fixed inputs and paths only.

Our evaluation has four objectives: 1) demonstrating that \projectname outperforms other frameworks with respect to both memory requirements  and execution latency (Section~\ref{sec:eva-overall}), 2) studying the performance effect of our key optimizations based on RDP (Section~\ref{sec:eva-breakdown}), 3)
further confirming the performance advantage of \projectname by evaluating it under  
different situations (Section~\ref{sec:eva-further}), and 4)
showing that \projectname performs well on different mobile platforms (i.e., \projectname has good portability).

\subsection{Evaluation Setup}
\label{sec:eva-setup} 

\noindent{\bf Dynamic Models and Datasets.} 
Our evaluation is conducted on three types of dynamic models: 1) models with shape dynamism, 2) models with control-flow dynamism, and 3) models with both shape and control-flow dynamism. The first category comprises 
five cutting-edge DNN models,  which are StableDiffusion~\cite{rombach2022high} (covering the Encoder part, referred to as SDE), SegmentAnything~\cite{kirillov2023segany}, Conformer~\cite{gulati2020conformer}, CodeBERT~\cite{feng2020codebert}, and YOLO-V6~\cite{li2022yolov6} (referred to as YL-V6). 
The second category includes DGNet~\cite{li2021dynamic}.
The third category consists of four models, including SkipNet~\cite{wang2018skipnet} (referred to as SNet), ConvNet-AIG~\cite{Veit2018} (referred to as CNet), RaNet~\cite{yang2020resolution}, and BlockDrop~\cite{wu2018blockdrop} (referred to as BDrop).

Table ~\ref{tab:eva_memory_report} characterizes these models by showing the nature of dynamism, target input types, model size, and the total number of layers.
Because the choice of training datasets has a negligible impact on the final inference latency or memory consumption (since the model size and structure are the same), this section reports results from one training dataset for each model. StableDiffusion-Encoder, SkipNet, DGNet, ConvNet-AIG, RaNet, and BlockDrop are trained on ImageNet dataset ~\cite{deng2009imagenet}; YOLO-V6 is trained on MS COCO dataset ~\cite{cocodataset}; SegmentAnything is trained on SA-1B dataset ~\cite{kirillov2023segany}; 
CodeBERT is pre-trained on ~\cite{codebertgithub}; and finally, 
Conformer is trained on Librispeech dataset ~\cite{Librispeech}. Since the model accuracy is the same across all frameworks, our evaluation focuses only on execution time and memory consumption. 

\noindent{\bf Test Samples and Setup.}
Our inference performance evaluation randomly selects 50 input samples from the corresponding validation dataset for each model. Specifically, for models that take images as input, i.e., 
YOLO-V6, SkipNet, ConvNet-AIG, RaNet, and BlockDrop, our evaluation randomly selects 50 input images from the ImageNet dataset, with the  size of  dimensions ranging from 224 to 640. DGNet does not support dynamic input shapes, but it does support dynamic control flow. Therefore, we only tested images with a dimension of 224 for DGNet. 
As YOLO-V6 only accepts images with dimensions that are multiples of 32, only a subset 
of inputs could be used. For StableDiffusion-Encoder and SegmentAnything,  the 50 
randomly selected input images have dimensions ranging from 64 to 224. 
For CodeBERT and Conformer, our evaluation randomly selects 50 input samples with sequential lengths ranging from 32 to 384.

The experiments are performed on a Samsung Galaxy S21 smartphone powered by a Snapdragon 888 processor~\cite{snapdragon888}. This processor features an octa-core Kryo 680 CPU, comprising one large core, three middle cores, and four small cores, and a Qualcomm Adreno 660 GPU with 1024 ALUs. %
Additionally, to demonstrate the portability of our approach, \projectname is also tested on an earlier generation of Snapdragon platform with more constrained resources, specifically the Snapdragon 835~\cite{snapdragon835} equipped with a Qualcomm Kryo 280 octa-core CPU, consisting of four middle cores and four small cores, and a Qualcomm Adreno 540 GPU with 384 ALUs. Our evaluation employs 8 threads on mobile CPUs and pipelined execution on mobile GPUs. The GPU execution uses a 16-bit floating-point representation, while the CPU execution uses a 32-bit floating-point representation. 
Each experiment is executed 50 times and only the average numbers are reported -- 
as the variance was negligible, it is not reported for readability.

\subsection{Overall Comparison}
\label{sec:eva-overall} 
This section focuses on the end-to-end memory reduction and execution latency gains of \projectname.

\noindent{\bf Overall Memory Consumption Comparison.}
Table~\ref{tab:eva_memory_report} presents a comparison of end-to-end memory consumption on a mobile CPU using \projectname, ONNX Runtime (ORT), MNN, and TVM with Nimble extension (TVM-N). As the results on mobile GPU show a similar trend, they are not included here. 
 `-'  implies that a model is not supported by a given framework. 
 The `Min' and `Max' columns indicate the minimum and maximum memory consumption (excluding the memory for holding the model itself because this part is the same for all frameworks).
 The last row of the table shows the geometric mean memory consumption of each framework normalized by \projectname. Its detailed calculation method is shown below the table and 
 is over the cases where execution is possible. 
 Among other frameworks, only MNN can support Conformer. SegmentAnything is  not supported by other frameworks as either certain key operators are missing, and/or there are limitations in optimization, leading to large model execution footprints. 
 
 Compared with other frameworks, \projectname has  significantly lower 
 memory consumption. Specifically, ORT, MNN, and TVM-N need to use  $3.64\times$, $1.37\times$, and $8.62\times$ memory, respectively,  over \projectname. 
 \projectname results in a greater reduction in memory consumption for image models (compared to other models) because image models generally have larger memory footprints, allowing for more significant optimization opportunities.
 It is worth noticing that TVM-N executes models as its own Android RPC application, which  
 is one of the causes of higher memory requirements. 

\begingroup
\setlength{\tabcolsep}{10pt}
\begin{table}[t!]
\centering
\caption{\revisioncr{Latency impact of input distribution on YOLO-V6. Each cell shows the latency speedup of \projectname over a corresponding baseline of ORT, MMN, or TVM-N.}}
\footnotesize{
\revisioncr{
\begin{tabular}{l|ccccc}
    \toprule
    Model & 1th & 25th & 50th & 75th & 100th \\
    \hline
    ORT	  & 1.43$\times$ & 1.66$\times$ & 1.95$\times$ & 2.33$\times$ & 2.52$\times$ \\
    MNN	  & 1.41$\times$ & 1.44$\times$ & 1.50$\times$ & 1.58$\times$ & 1.65$\times$ \\
    TVM-N & 2.13$\times$ & 2.52$\times$ & 3.03$\times$ & 3.67$\times$ & 3.90$\times$ \\
    \bottomrule
\end{tabular}
}
}
\label{tab:eva_speedup_distribution}
\end{table}
\endgroup

\noindent{\bf Overall Latency Comparison.}
Table ~\ref{tab:eva_performance_report} presents a comparison of end-to-end latency for \projectname against other frameworks on both mobile CPU and GPU. The table includes the minimum and maximum latency observed across different input samples for each model. On mobile CPU, \projectname achieves an average speedup of $2.5\times$, $1.7\times$, and $2.7\times$ compared to ONNX Runtime, MNN, and
TVM-N, respectively. TVM-N does not support dynamic models on a  mobile GPU.  
Compared against the other two frameworks on mobile GPU, \projectname achieves a speedup of $3.9\times$ and $2.3\times$ over ORT and MNN, respectively. Notably, the minimum latency achieved by \projectname on mobile GPU is significantly lower than other frameworks for ConvNet-AIG, RaNet, and BlockDrop models. This is because our optimizations can handle different cases and mitigate the effect of execution path variations.
\revisioncr{
It is worth pointing out that the distribution of inputs could impact results. However, it does not change our conclusion. To show this impact more explicitly, we conduct a set of experiments on YOLO-V6 by selecting 50 input samples from different percentiles ranging from 1st to 100th, and our results are as shown in Table ~\ref{tab:eva_speedup_distribution}.
}

\begin{figure}[t]
    \centering
        \subfloat[Mobile CPU]{
            \includegraphics[width=0.98\columnwidth]{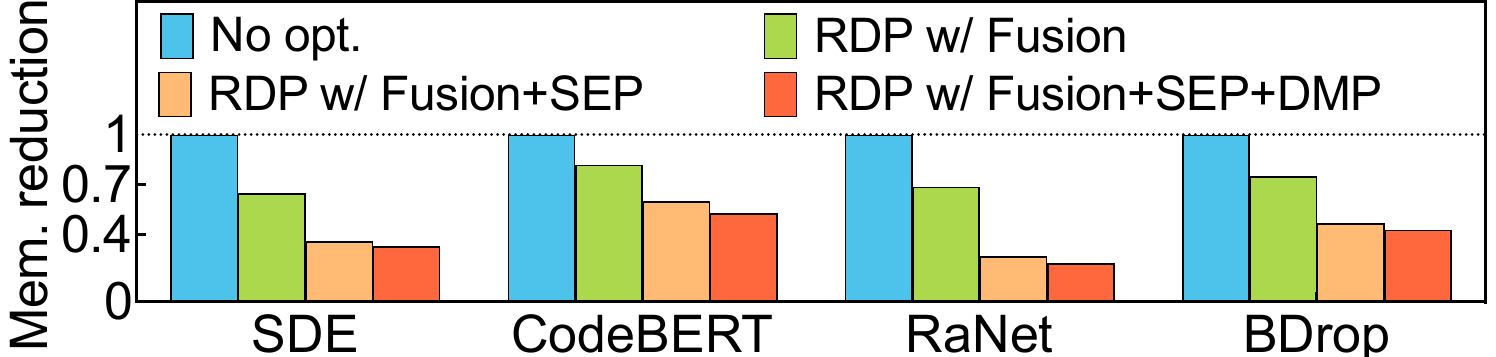}
        }
        \caption{\textbf{Memory reduction of different optimizations on CPU.} Over the baseline w/o any RDP-enabled optimization (No opt.) 
        }
        \vspace{-0.9em}
    \label{fig:eva_memory_breakdown}
\end{figure}

\begin{figure}[t]
\centering
\subfloat[Mobile CPU]{
    \includegraphics[width=0.98\columnwidth]{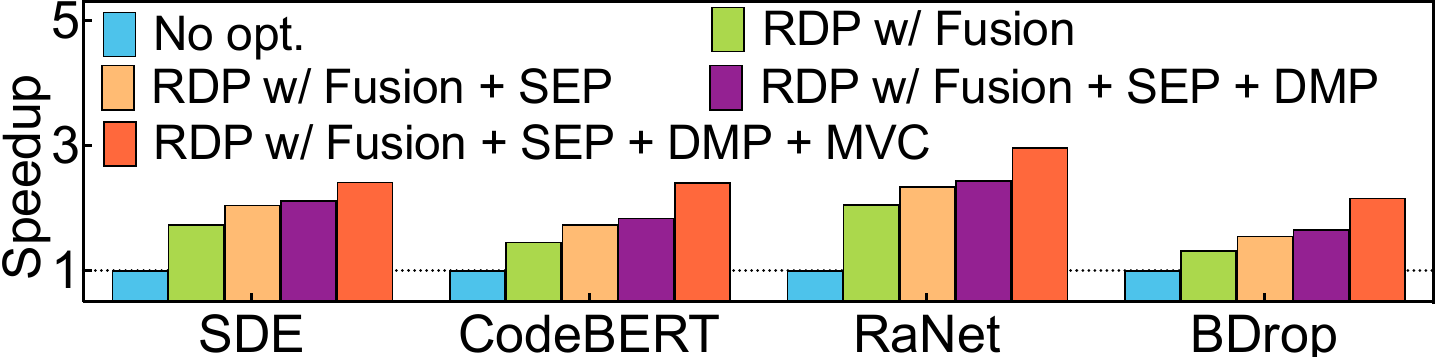}
}\\
\vspace{-1em}
\subfloat[Mobile GPU]{
    \includegraphics[width=0.98\columnwidth]{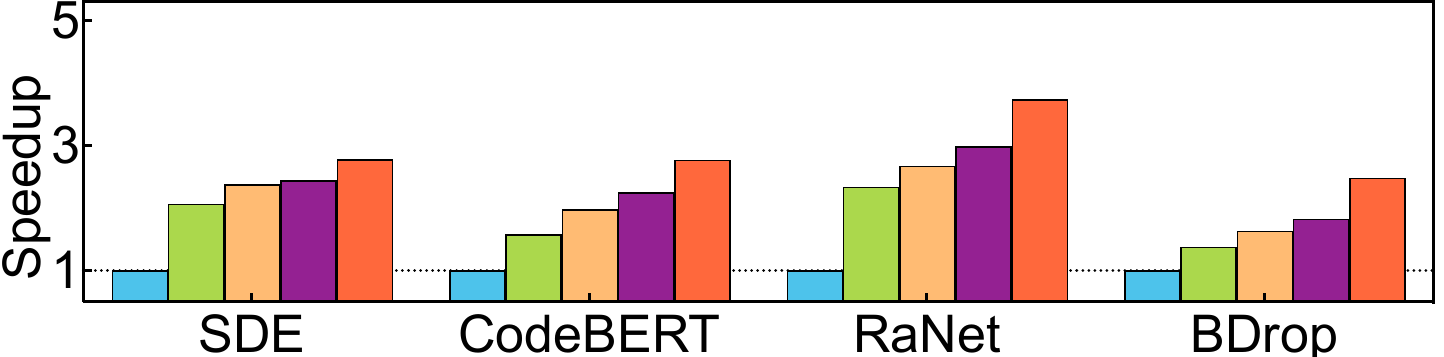}
}
\caption{\textbf{Execution speedup of different opt. on CPU and GPU.} Over the baseline w/o any RDP-enabled optimization (No opt.)  
}
\label{fig:eva_performance_breakdown}
\vspace{-1em}
\end{figure}

\subsection{Optimization Breakdown Analysis}
\label{sec:eva-breakdown} 
This section studies the individual impact  of the  key optimizations in \projectname on  both memory consumption and latency.

\noindent{\bf Memory Reduction w/ Different Optimizations.}
Figure ~\ref{fig:eva_memory_breakdown}  evaluates the memory reduction achieved  through different optimizations for 4 models (StableDiffusion-Encoder, CodeBERT, RaNet, and BlockDrop), including RDP-enabled operator fusion ({\tt Fusion}), static execution planning ({\tt SEP}), and dynamic memory planning ({\tt DMP}). The results for other models exhibit a similar trend and are excluded due to space limitations. The baseline version is referred to as {\tt No opt} -- despite the name, it includes  general static optimizations, such as static operator fusion and constant folding. Building on this version, we study the benefits of optimizations enabled by RDP analysis. On mobile CPU, operator fusion, static execution planning, and dynamic memory planning bring $18\%$ to $30\%$, an extra $22\%$ to $37\%$, and another extra $3\%$ to $7\%$ memory reduction, respectively.
Multi-version code generation ({\tt MVC}) is primarily designed for latency improvement, its impact on memory reduction is negligible. 
The memory reduction on mobile GPU is omitted because our optimizations are general to both CPU and GPU, and the results are similar.

\noindent{\bf Latency Reduction w/ Different Optimizations.}
Figure~\ref{fig:eva_performance_breakdown} presents the speedup breakdown of our key optimizations on the same 4 models. %
On mobile CPU, our RDP-based operator fusion yields $1.3\times$ to $1.9\times$ speedup compared to {\tt No opt}. Additionally, static execution planning provides $1.1\times$ to $1.3\times$ speedup, and dynamic memory planning gains $1.04\times$ to $1.1\times$ speedup, and Multi-version code generation brings an extra $1.3\times$ to $1.6\times$ speedup. On mobile GPU, these numbers are $1.4\times$ to $2.3\times$, $1.2\times$ to $1.3\times$, $1.06\times$ to $1.2\times$, and $1.4\times$ to $1.7\times$, respectively.
Our optimizations provide more benefits for mobile GPU since GPU is more sensitive to memory and 
data movement and supports a higher 
degree of parallelism. 
We further study each optimization with more profiling results.

\begin{figure}[t]
\centering
    \subfloat[Layer count]{
        \includegraphics[width=0.48\columnwidth]{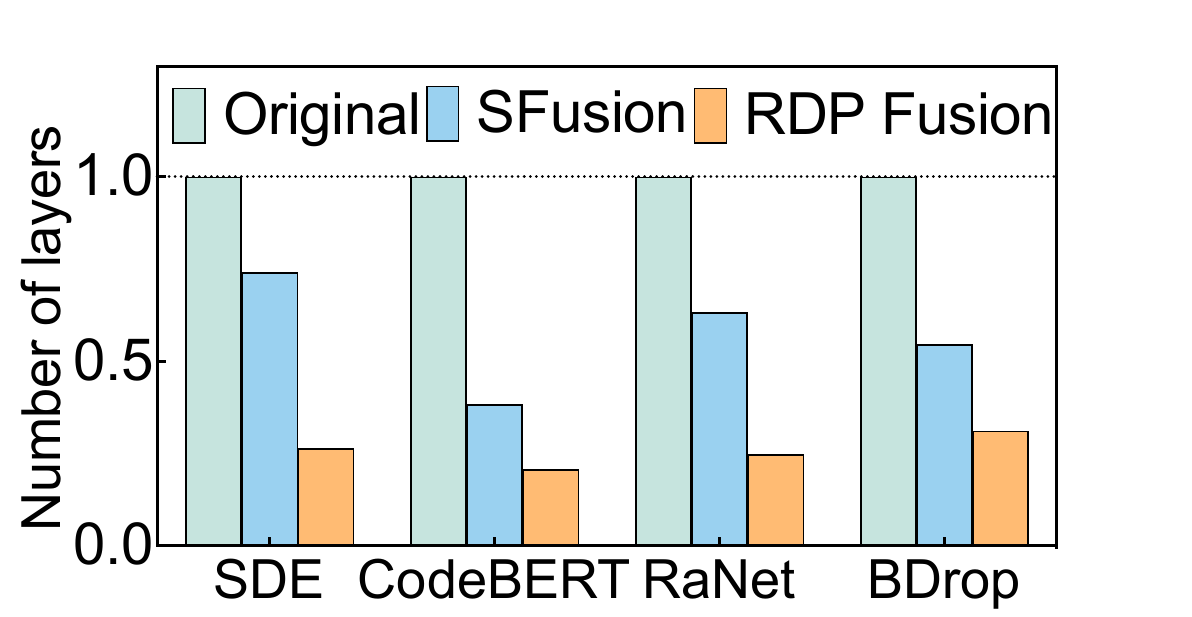}
    }
    \subfloat[IR size]{
        \includegraphics[width=0.48\columnwidth]{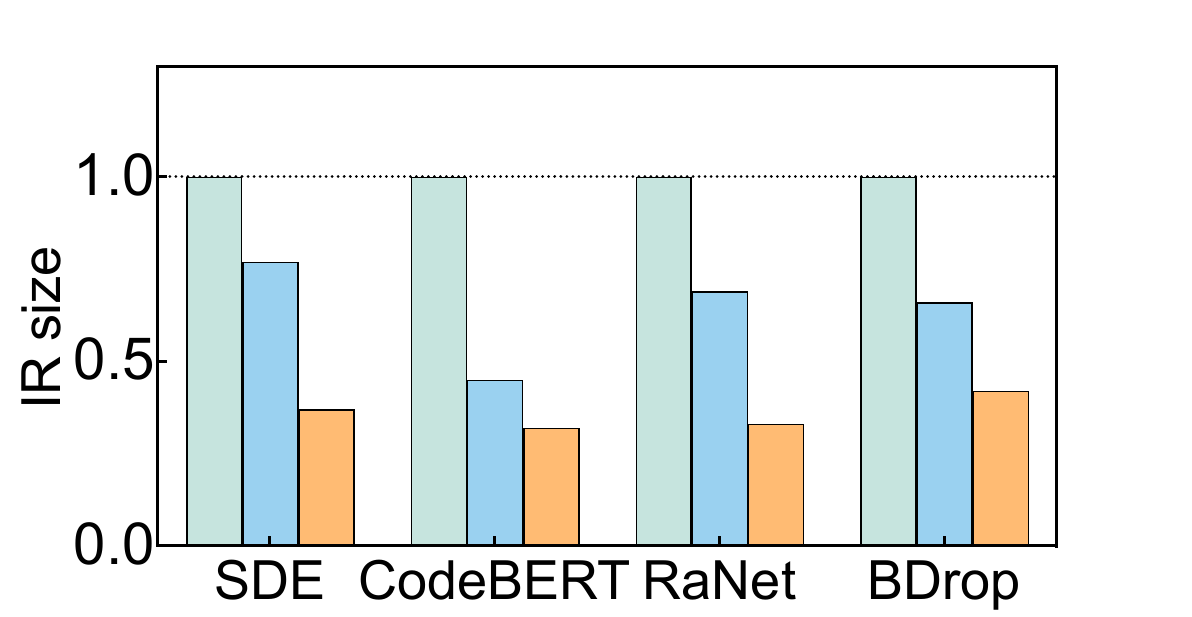}
    }
    \caption{{\bf Further break down effect of existing static fusion (SFusion) and RDP-based fusion (RDP Fusion).} For both layer count and intermediate result size, normalized by no fusion opt.}
    \vspace{-1.5em}
\label{fig:eva_performance_fusion}
\end{figure}

\begin{figure}[t]
\centering
    \subfloat[Sub-graph percentage]{
        \includegraphics[width=0.48\columnwidth]{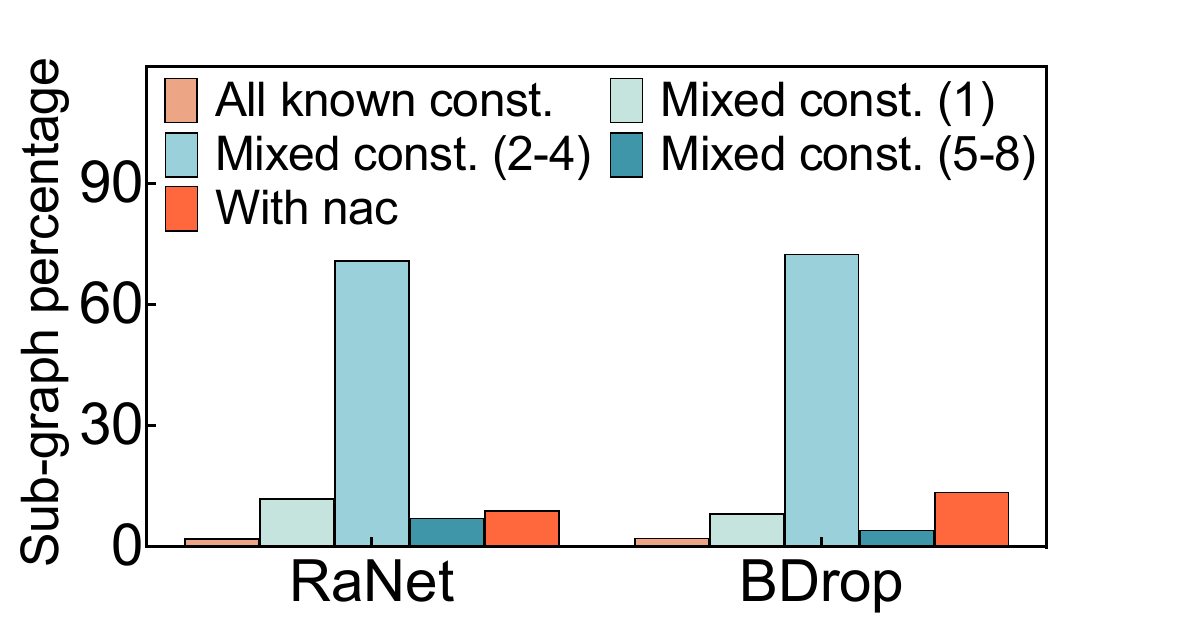}
    }
    \subfloat[Latency percentage]{
        \includegraphics[width=0.48\columnwidth]{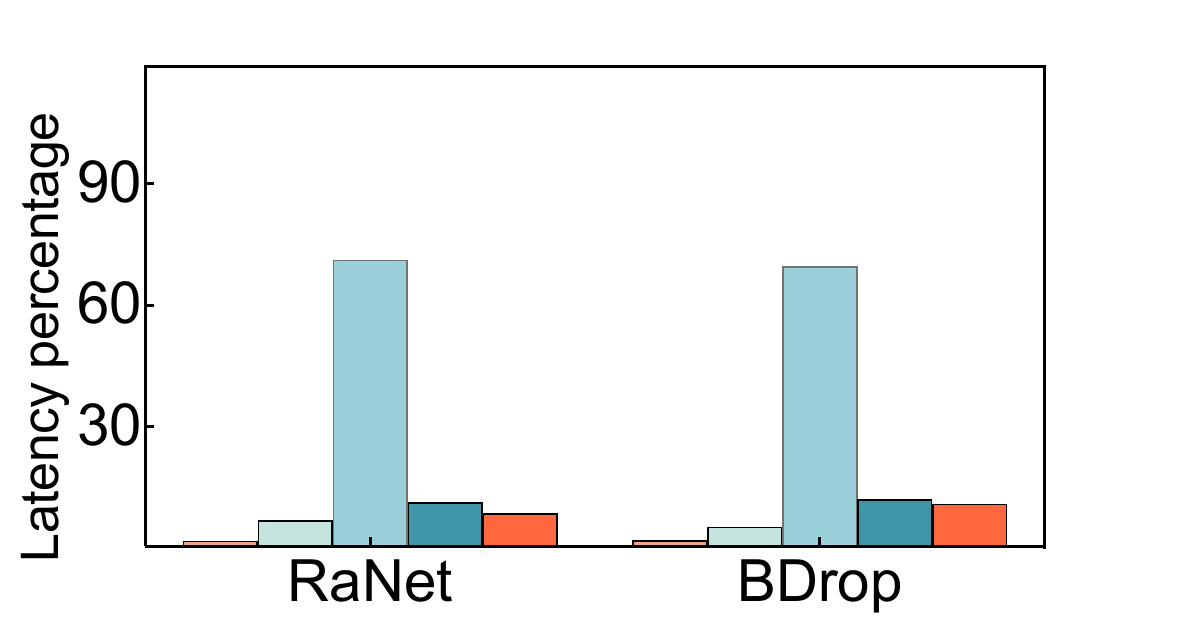}
    }
    \caption{{\bf The percentage of different types of sub-graph.}}
\label{fig:eva_performance_profiling_execution}
\end{figure}

\noindent{\bf RDP-enabled Operator Fusion.}
Figure~\ref{fig:eva_performance_fusion} 
further breaks down the effect of existing operator fusion for static DNNs only ({\tt SFusion}) and our RDP-enabled operator fusion ({\tt RDP Fusion}) on these four dynamic DNNs. These results are normalized by the original DNN without fusion ({\tt Original}). 
{\tt SFusion} reduces the layer counts by $26\%$ to $61\%$; while {\tt RDP Fusion} further reduces the layer counts by $16\%$ to $46\%$ additionally by leveraging RDP analysis results. In terms of intermediate result (IR) size, {\tt RDP Fusion} saves an additional $13\%$ to $40\%$ on top of {\tt SFusion}.

\noindent{\bf Subgraph Data.} To better understand execution and memory planning, this part studies how many sub-graphs can benefit from RDP analysis results.
Figure ~\ref{fig:eva_performance_profiling_execution} (a) shows the percentage of different sub-graphs, i.e. those with all known constant shapes, with mixed constant shapes, and with statically unknown ($nac$) only  for 2 representative models. The numbers (1, 2-4, and 5-8) after {\tt Mixed const} denote the number of code versions that are required to optimize this sub-graph (the lower the better). This result shows that over 90\% of the sub-graphs belong to all known constant or mixed constant categories whose execution plan and memory plan can be optimized by our framework. To further confirm this, Figure ~\ref{fig:eva_performance_profiling_execution} (b) shows the latency percentage of each kind of sub-graphs.

\begin{figure}[t]
\centering
    \subfloat[Inference time]{
        \includegraphics[width=0.48\columnwidth]{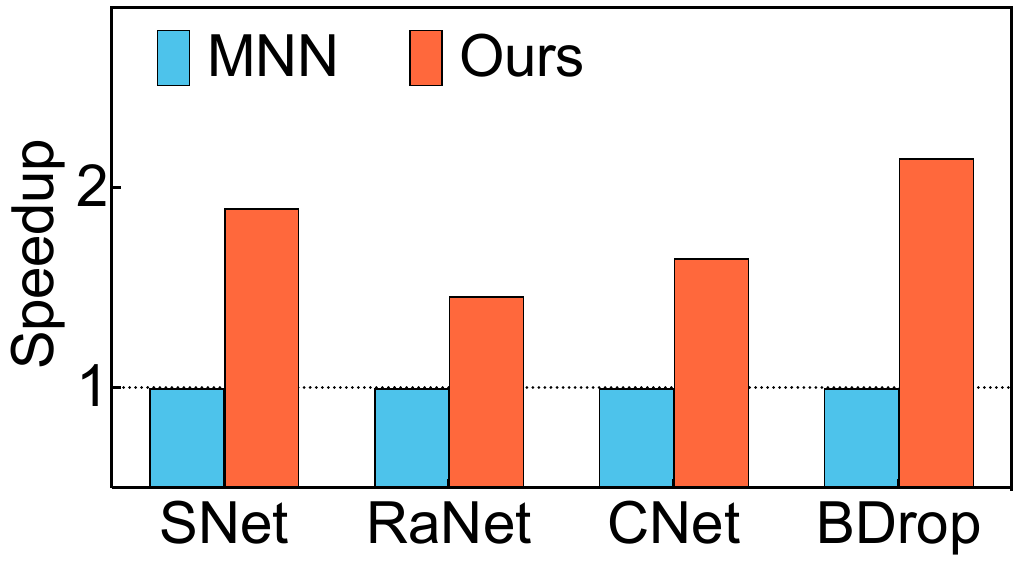}
    }
    \subfloat[Memory consumption]{
        \includegraphics[width=0.48\columnwidth]{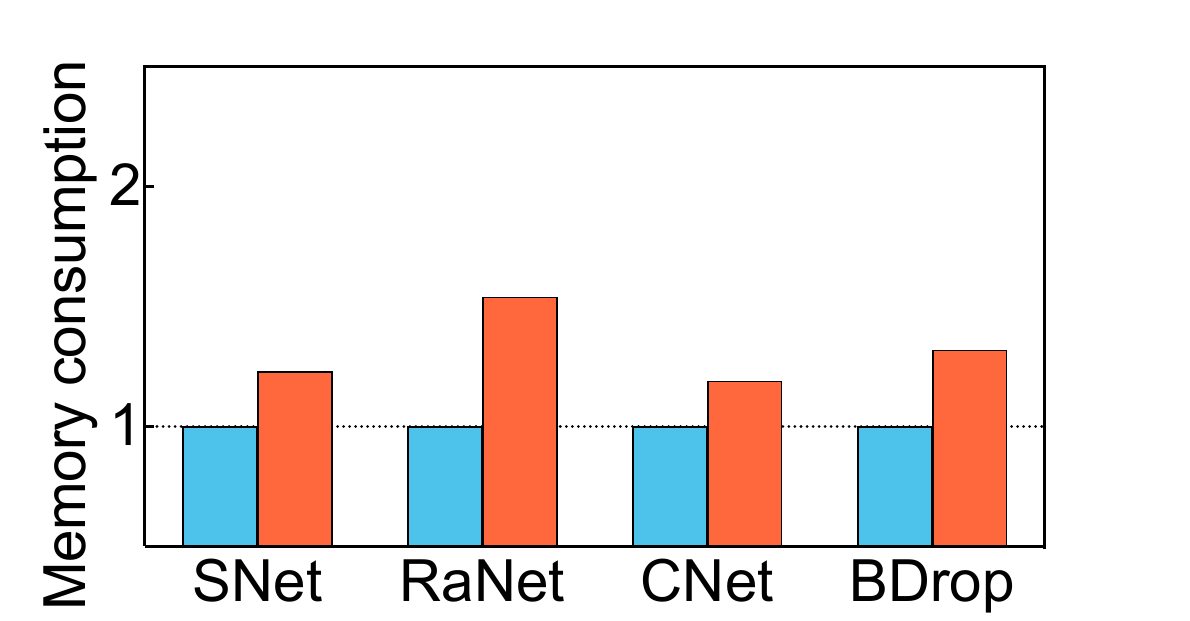}
    }
    \caption{\textbf{Latency and memory consumption comparison between \projectname and MNN with the same execution path.}}
    \vspace{-1.5em}
\label{fig:eva_performance_same_path}
\end{figure}

\subsection{Further Performance Analysis}
\label{sec:eva-further} 
This section further studies \projectname under different cases.

\noindent{\bf Latency Comparison with the Same Execution Path.}
To provide an apple-to-apple comparison for control-flow dynamism, this test disables the control-flow logic 
in 4  models (SkipNet, RaNet, ConvNet-AIG, and BlockDrop) that have 
control-flow dynamism. Our execution included all paths, including all branches in the {\em <Switch, Combine>} pairs.  
Figure ~\ref{fig:eva_performance_same_path} illustrates the performance comparison with MNN because MNN performs the best among all baseline frameworks we compared. \projectname achieves $1.5\times$ to $2.0\times$ speedup and $1.2\times$ to $1.5\times$ memory reduction on the mobile CPU. This result further validates the effect of our RDP analysis and fusion, execution, and memory optimizations based on it even without the dynamic branch selection capability of \projectname. 

\begin{figure}[t]
\centering
    \subfloat[Mobile CPU]{
        \includegraphics[width=0.48\columnwidth]{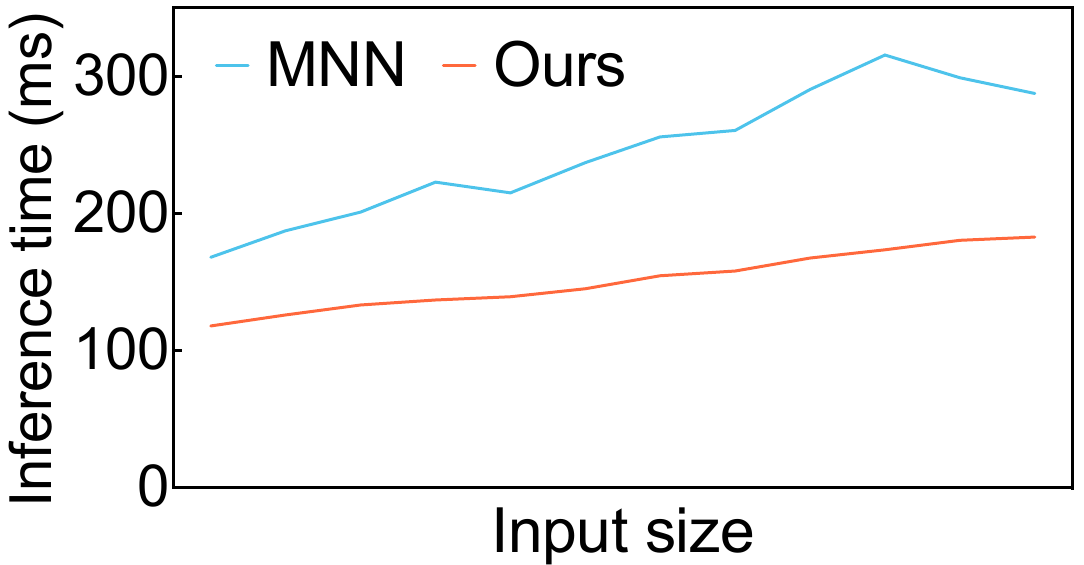}
    }
    \subfloat[Mobile GPU]{
        \includegraphics[width=0.48\columnwidth]{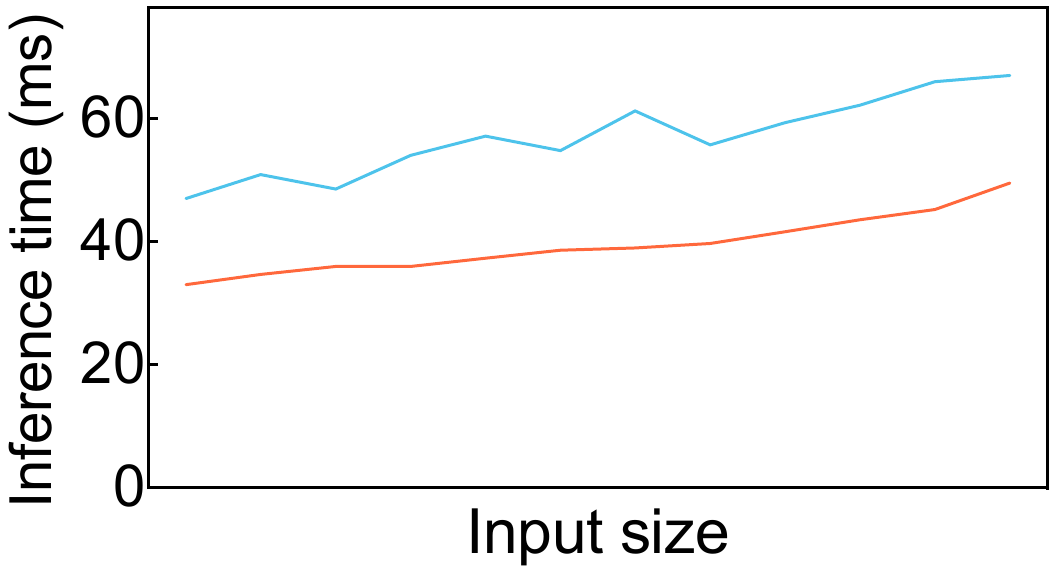}
    }
    \caption{\textbf{Performance variation with different input sizes  (shapes).} The data is collected from YOLO-V6. A larger input size means more computations.}
\label{fig:eva_performance_variation}
\end{figure}

\noindent{\bf Latency Comparison with Different Input Sizes.}
To demonstrate the stability of \projectname, this test randomly selects 15 input shapes for YOLO-V6, and Figure ~\ref{fig:eva_performance_variation}, shows their inference latency with MNN and \projectname. 
These results demonstrate that \projectname outperforms MNN in terms of both latency and stability across increasing input sizes on mobile CPUs and GPUs. Specifically, \projectname exhibits lower and more consistent latency, while MNN exhibits significant variations.

\begin{figure}[t]
    \centering
        \subfloat[Mobile CPU]{
            \includegraphics[width=0.48\columnwidth]{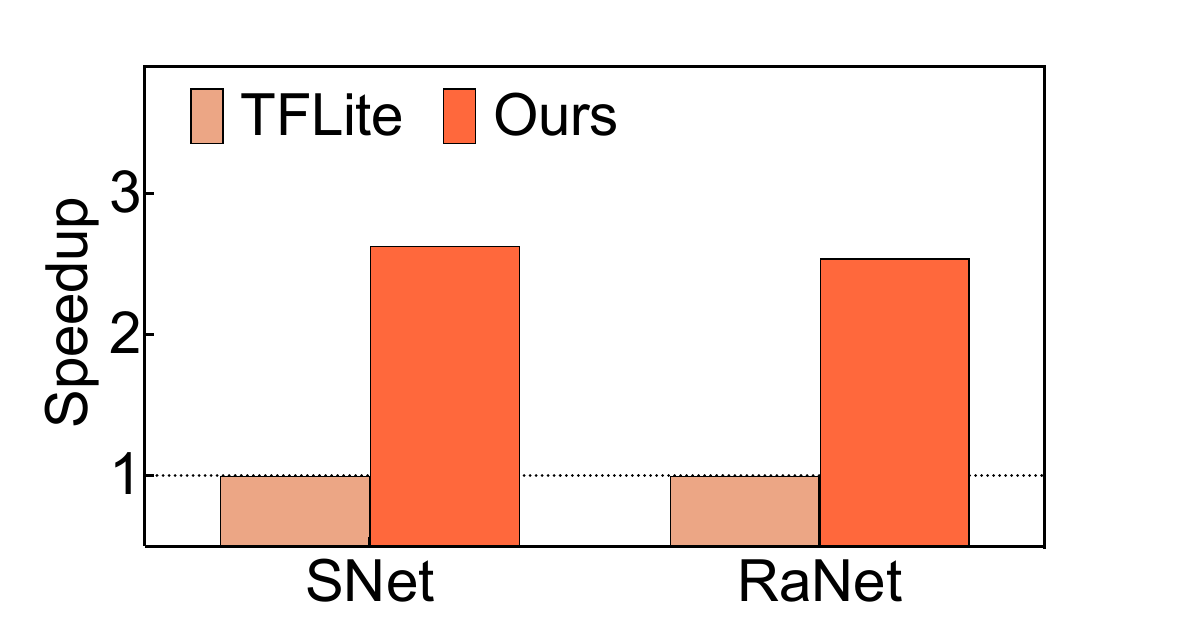}
        }
        \subfloat[Mobile GPU]{
            \includegraphics[width=0.48\columnwidth]{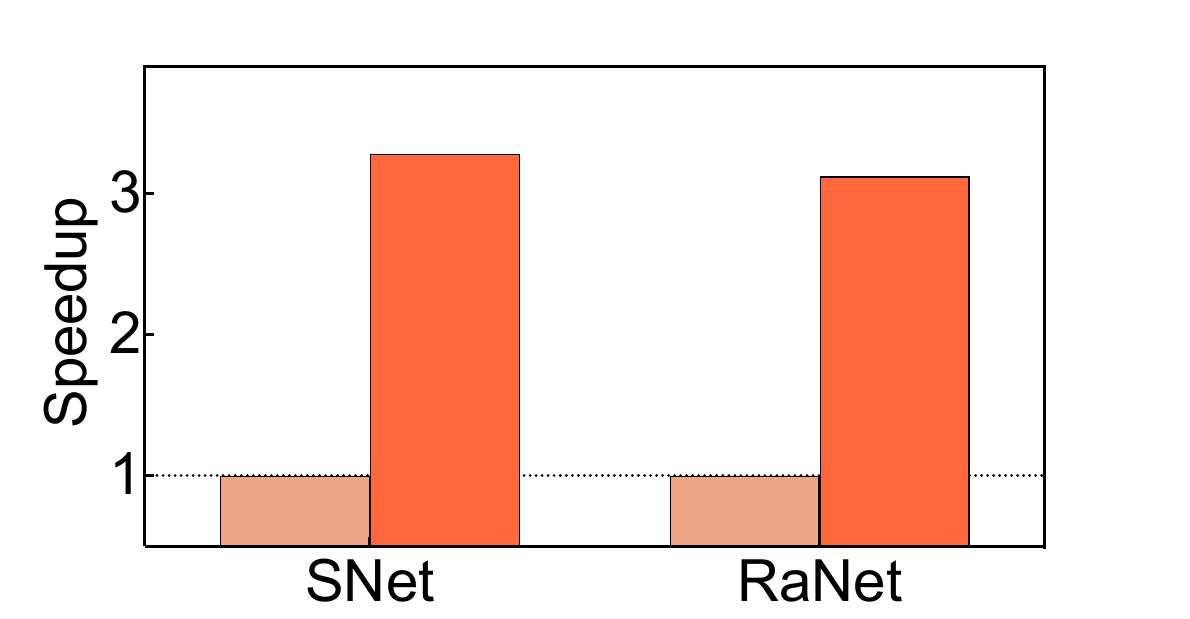}
        }
        \caption{\textbf{Speedup with the same memory consumption.} }
    \label{fig:eva_memory_budget}
\end{figure}

\noindent{\bf Latency Comparison with Fixed Memory Budget.}
Figure~\ref{fig:eva_memory_budget} presents a latency comparison between \projectname and TFLite with the same memory budget. %
Specifically, TFLite fixes its memory consumption to match \projectname's, and uses the XLA rematerialization policy~\cite{tensorflow-xla} to handle the out-of-memory cases. \projectname outperforms TFLite by an even greater margin.
Additionally, \projectname demonstrates a higher speedup on mobile GPU compared to mobile CPU due to the longer time required for mobile GPU to materialize intermediate tensors from its cache into main memory because of memory mapping. 

\begin{figure}[t]
    \centering
        \subfloat[Mobile CPU]{
            \includegraphics[width=0.48\columnwidth]{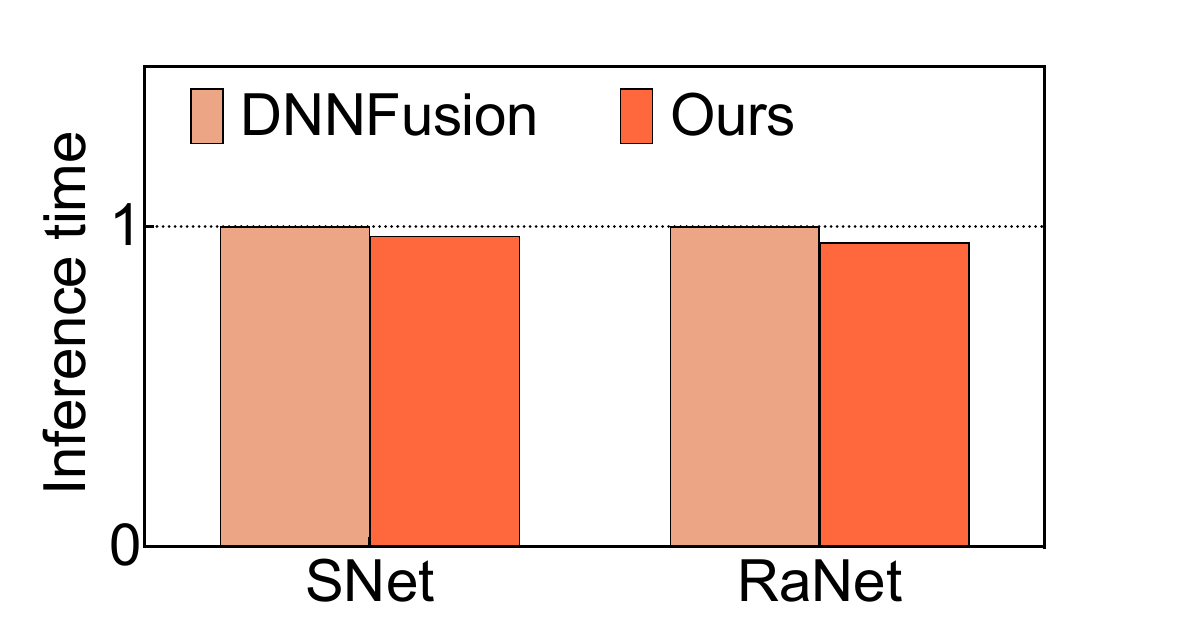}
        }
        \subfloat[Mobile GPU]{
            \includegraphics[width=0.48\columnwidth]{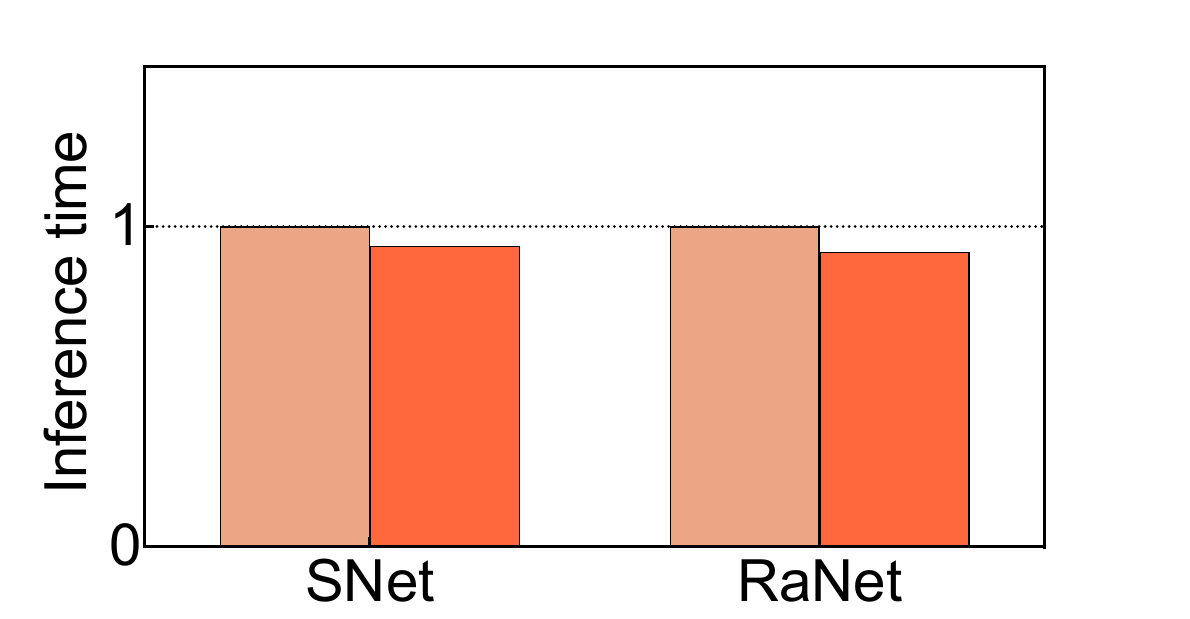}
        }
        \caption{Inference time comparison with DNNFusion for static models (i.e., with both frozen shapes and control flows).}
    \label{fig:eva_static_model}
\end{figure}

\noindent{\bf Latency Comparison with Static Models.}
\revisioncr{
Figure~\ref{fig:eva_static_model} examines the latency overhead of \projectname in contrast to our baseline, DNNFusion~\cite{niu2021dnnfusion}, for static models. 
Specifically, we evaluate the latency in SkipNet and RaNet where dynamic values were fixed statically and fully propagated, ensuring the absence of unknown values and dynamic control flows. As shown in Figure~\ref{fig:eva_static_model}, \projectname incurs an average overhead of 3\% and 7\% performance slowdown when compared to the completely optimized static DNNFusion. This is attributed to the fact that DNNFusion, with full information available, results in a more comprehensive fusion optimization and does not include dynamic memory planning overhead.
}

\subsection{Portability}
\begin{figure}[t]
    \centering
        \subfloat[Mobile CPU]{
            \includegraphics[width=0.48\columnwidth]{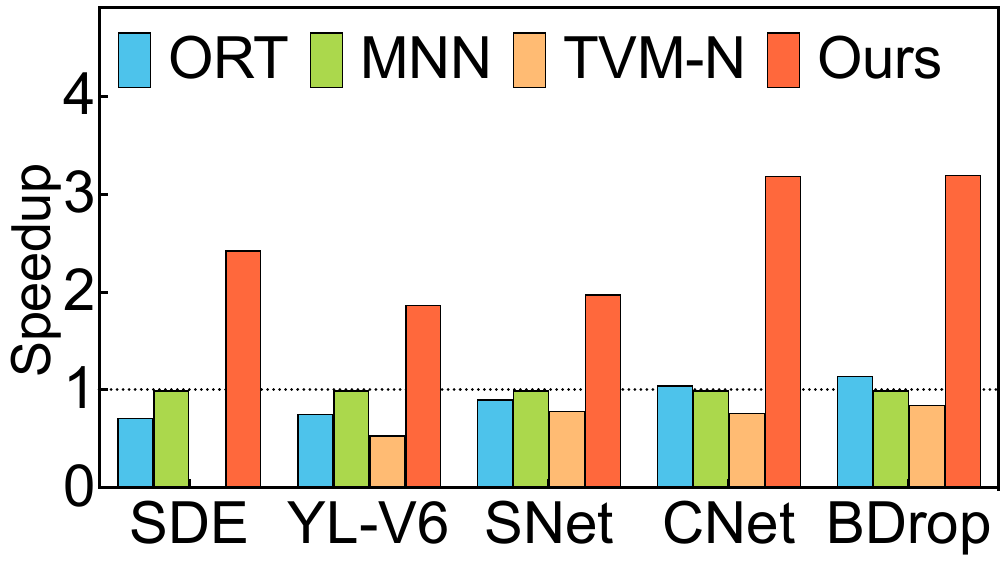}
        }
        \subfloat[Mobile GPU]{
            \includegraphics[width=0.48\columnwidth]{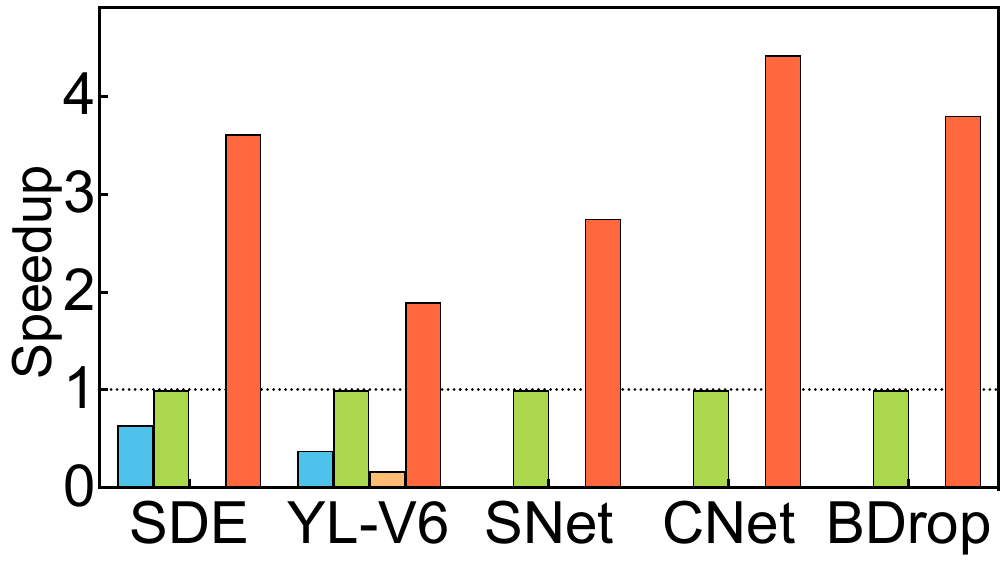}
        }
        \caption{\textbf{Portability evaluation.} The results are collected on Snapdragon 835. An empty bar means the model is not supported by the framework. Results are normalized by MNN for readability.}
    \label{fig:eva_portibility}
\end{figure}

\revisioncr{
\label{sec:eva-portability} 
To further investigate the effectiveness of portability, Figure ~\ref{fig:eva_portibility} shows the execution speedup of \projectname over other frameworks on another  mobile 
device -- Snapdragon 835, and 5 models (StableDiffusion-Encoder, YOLO-V6, SkipNet, ConvNet-AIG, and BlockDrop). \projectname achieves similar speedup trends, and interestingly, it achieves higher speedups on this earlier generation of SoC because this SoC has more restricted resources (e.g., cache size and memory throughput). The RDP-based optimizations employed in \projectname significantly reduce memory requirements, leading to improved performance on these platforms.
}

\section{Related Work}

\noindent{\bf Dynamic Neural Network Optimizations.}  
Type analysis and type inference~\cite{milner1978theory,harper1995compiling,crary1999flexible,siek2007gradual,lattner2021mlir} are widely used to analyze tensor shapes, thus assisting  in 
Dynamic Neural Network optimizations. 
Nimble~\cite{shen2021nimble}, which has been integrated into TVM, is a compilation-based Dynamic DNN framework. This framework relies on expensive dynamic functions to interpret dynamic shapes at the  runtime. This implementation, which we have extensively compared against, limits the opportunities for optimized code generation, such as performing operator fusion.
DISC~\cite{zhu2021disc} extends MLIR-HLO~\cite{lattner2021mlir}  and  propagates the shape information for operators that have  certain  constraints, e.g. same dimensions ( 
the case of {\tt Activation}) and same size (the case of {\tt Transpose}). \projectname provides a more comprehensive operator classification based on dynamism degrees, bringing in significantly enhanced optimization opportunities.
Axon~\cite{collins2022axon} is a programming language that allows  specification 
of symbolic shapes for input and output tensors for computational graphs. It uses a constraint solver to find shapes whereas \projectname uses a forward and backward data-flow analysis (RDP), which also alleviates additional 
programmer involvement. 
In addition, \projectname includes a set of opts enabled by RDP.

Less closely related to \projectname,  DietCode~\cite{zheng2022dietcode} proposes an auto-scheduler framework based on TVM for dynamic shapes. The framework builds a cost model to predict runtime performance and reduces the search space to find optimal runtime parameters (e.g., loop tiling). 
Cortex~\cite{fegade2021cortex}, Cavs~\cite{xu2018cavs}, and another effort~\cite{jeong2018improving} mainly aim to address recursive dynamism of neural networks, different from \projectname's focus.
Other efforts focus on dynamic batching for inference~\cite{looks2017deep,gao2018low,fang2021turbotransformers, zhai2022bytetransformer} or are designed for dynamic DNN training~\cite{neubig2017dynet}. %

\noindent{\bf DNN Execution and Memory Optimizations.} 
Several studies exist for operator execution order scheduling, such as~\cite{liberis2019neural,lin2020mcunet,ahn2020ordering}. Among these efforts~\cite{liberis2019neural, lin2020mcunet} focus on minimizing peak memory consumption by reordering operators for resource-constrained devices (e.g., MCUs), and effort~\cite{ahn2020ordering} proposes an optimized scheduling framework for complex models (irregularly wired neural networks). These approaches rely on static shapes only. 
There have aldo  been recent efforts on optimizing memory allocation planning and memory management for DNNs. Works such as~\cite{levental2022memory,pisarchyk2020efficient} have designed various heuristic memory planning algorithms for static DNNs only. 
TelaMalloc~\cite{maas2022telamalloc} performs memory management on the fly for static control-flow graphs with known intermediate tensor shapes and sizes. It does not fully consider the DNN control-flow dynamism and dynamic shapes. 
A possible future work can be to integrate  our RDP analysis and TelaMalloc's combination 
of 
heuristics with a solver-based approach  to further improve our memory planning. 
When the available memory is limited, 
rematerialization~\cite{jain2020checkmate, kirisame2020dynamic} and recomputation~\cite{bulo2018place}  
methods achieve a trade-off between  memory consumption and execution latency. 
These aspects can be considered for dynamic DNNs in the future. 

\noindent{\bf DNN Inference Engines on Mobile.} 
Support for DNN  inference  on mobile devices has become an area of active research in recent years.  Efforts such as MCDNN~\cite{han2016mcdnn}, DeepX~\cite{lane2016deepx}, DeepMon~\cite{huynh2017deepmon}, DeepSense~\cite{yao2017deepsense}, and DeepCache~\cite{xu2018deepcache} have primarily concentrated on optimizing the execution of static DNNs with static shapes and control flow.  
TensorFlow Lite (TFLite)~\cite{TensorFlow-Lite}, Pytorch-Mobile ~\cite{paszke2019pytorch}, TVM~\cite{chen2018tvm}, and MNN~\cite{Ali-MNN} provide support for dynamic shapes  relying on reinitialization or conservative (maximum) memory allocation. They either do not support dynamic control flow or require executions of all paths with a stripping of invalid results. 
As shown in our evaluation, these methods introduce high runtime overhead.  
One of the previous systems for static DNNs, DNNFusion~\cite{niu2021dnnfusion}, 
also involved a classification of DNN operators, however, the 
classification introduced here is orthogonal.

\section{Discussion and Future Work}

\noindent\revisioncr{\textbf{Generalizing to Other Platforms.}
The proposed techniques, such as RDP analysis, RDP-enabled fusion, and execution and memory planning, have broad applicability to various platforms, including data-center GPUs. This is particularly true for single-input inference scenarios. 
One potential nuance that may arise is the distinction between data-center GPUs and mobile GPUs in terms of their ability to perform batched inference. 
Unlike mobile GPUs, data-center GPUs have the capacity to process multiple inputs concurrently, thereby maximizing their computational power.
However, it is possible that different input samples within a batch may necessitate the use of different execution paths.
Therefore, the integration of dynamic batching with dynamic neural networks presents a potential direction for future research. 
}

\noindent\revisioncr{
\textbf{Scalability of Handling LLMs.} The optimizations in \projectname can also be applied to massive large language models (LLMs). 
One of the primary procedures we employ is graph partitioning, as elaborated in Section ~\ref{subsec:sep}. This procedure involves dividing the entire computational graph into a collection of sub-graphs, each of which encompasses a restricted number of layers. 
The optimal solution is determined offline for each sub-graph. 
However, Language Models (LLMs) have been characterized by an incredibly large number of parameters, numbering in the billions ~\cite{zeng2022glm,DatabricksBlog2023DollyV2,touvron2023llama}.
This poses a significant challenge for mobile devices in terms of computation and resource requirements. 
Our future work will enhance \projectname by combining it with the model pruning and quantization advances ~\cite{niu2020patdnn,NEURIPS2022_80133d0f,jin2022f8net} to achieve an even better performance.
}

\noindent\revisioncr{
\textbf{Extending beyond ONNX.}
Operator classification and associated optimization designs are also not limited to ONNX or other inference formats (e.g., TFLite, Caffe2). 
This is because our proposed analysis is based on the degree of dynamism defined by the computation logic of an operator and the relationship between its input and output, rather than relying on the specific representation or format of the operator. 
Some formats have yet to fully support dynamic computational graphs.
For instance, PyTorch supports exporting models with dynamic shapes (such as Input Shape Determined Output, Input Shape Determined Output Shape, and Input Shape \& Value Determined Output Shape) to ONNX. However, it is unable to convert models with dynamic control flow to ONNX. To address this limitation, we added a customized ONNX operator pair <Switch, Combine> (as shown in Figure ~\ref{fig:dynamism_examples}d) and registered a customized export routine on PyTorch specifically for models with a dynamic control flow. 
\projectname does have limitations in handling very complicated (or user-defined) dynamic models (such as Graph Neural Networks or DNNs involving recursive executions) that can be represented well in PyTorch. We leave this further optimization as a future work.
}

\section{Conclusions}\label{sec:conclusion} 

This paper has presented a comprehensive framework, \projectname, for optimizing 
DNNs.  \projectname classifies common operators of Dynamic DNNs into four types, 
and comprises  a novel static dataflow analysis (RDP). This is 
followed by  a set of optimizations enabled by RDP for Dynamic DNNs, including operator fusion, static execution (order) planning, dynamic memory allocation planning, and multi-version code generation.  \projectname is extensively evaluated on  a mobile 
system with 10 emerging dynamic DNNs and the evaluation results show  that it saves up to $88\%$ memory consumption and brings up to $3.9\times$ execution speedup over four state-of-the-art DNN execution frameworks.  As the underlying 
techniques are general and applicable to other devices as well,  our future work  
will evaluate \projectname's efficacy on other  devices (e.g., 
edge GPUs and  Raspberry Pi). 

\begin{acks}
\revisioncr{
The authors would like to express their gratitude to the anonymous reviewers and shepherd for their insightful and detailed comments. 
All of these have significantly contributed to the enhancement of this paper. 
This work was supported in part by the National Science Foundation (NSF) under the awards of CCF-2047516 (CAREER), CCF-2146873, CCF-2333895, CCF-2334273,  CNS-2230944, CNS-2341378,  IIS-2142681, III-2008557, and OAC-2333899.
Any errors and opinions are not those of the NSF and are attributable solely to the author(s). 
The authors also acknowledge William \& Mary Research Computing for providing computational resources.
}
\end{acks}

\balance
\bibliographystyle{ACM-Reference-Format}
\bibliography{reference}

\end{document}